%% file: main.tex
\documentclass[10pt,twocolumn,letterpaper]{article}

\usepackage{cvpr}              


\definecolor{cvprblue}{rgb}{0.21,0.49,0.74}
\usepackage[pagebackref,breaklinks,colorlinks,allcolors=cvprblue]{hyperref}
\usepackage{hyperref}
\usepackage{url}
\usepackage[accsupp]{axessibility}
\usepackage{graphicx}
\usepackage{amsmath}
\usepackage{amssymb}
\usepackage{booktabs}
\usepackage{enumitem}
\usepackage{multirow}
\usepackage{wrapfig}
\usepackage{makecell}
\usepackage[accsupp]{axessibility}
\usepackage{footnote}
\usepackage[para]{footmisc}

\usepackage{comment}
\usepackage{xcolor}
\usepackage{soul, color}

\usepackage{algorithm}
\usepackage[noend]{algcompatible}
\usepackage{arydshln} 
\usepackage[
  separate-uncertainty = true,
  multi-part-units = repeat
]{siunitx}

\usepackage[capitalize]{cleveref}
\crefname{section}{Sec.}{Secs.}
\Crefname{section}{Section}{Sections}
\Crefname{table}{Table}{Tables}
\crefname{table}{Tab.}{Tabs.}

\def\paperID{10277} 
\def\confName{CVPR}
\def\confYear{2025}
\newcommand{\nickname}{LogoSP}

\title{\nickname{}: Local-global Grouping of Superpoints for Unsupervised Semantic Segmentation of 3D Point Clouds}

\begin{document}

\author{Zihui Zhang \textsuperscript{1,2}\footnotemark[2] \quad  Weisheng Dai \textsuperscript{1,2}\footnotemark[2] \quad  Hongtao Wen \textsuperscript{1,2} \quad  Bo Yang \textsuperscript{1,2}\footnotemark[1] \\
 \textsuperscript{1} Shenzhen Research Institute, The Hong Kong Polytechnic University \\
 \textsuperscript{2} vLAR Group, The Hong Kong Polytechnic University\\
{\tt\small zihui.zhang@connect.polyu.hk, bo.yang@polyu.edu.hk}}

\twocolumn[{%
\renewcommand\twocolumn[1][]{#1}%
    \maketitle
    \begin{center}
        \vspace{-20pt}
        \centering
        \includegraphics[scale=0.48]{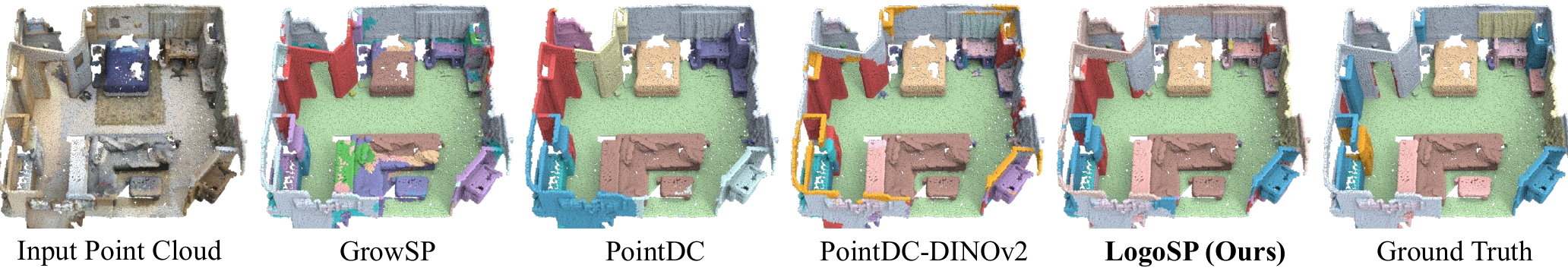}
        \vspace{-5pt}
        \captionof{figure}{Given an input point cloud with complex structures from ScanNet dataset, our \nickname{} automatically discovers accurate semantic classes such as \textit{sofa} and \textit{table} without needing human labels in training, clearly outperforming all existing baselines.}
        \label{fig:opening}
        \vspace{5pt}
    \end{center}
}]

\renewcommand{\thefootnote}{\fnsymbol{footnote}}
\footnotetext[1]{ Corresponding author}
\footnotetext[2]{ Equal contribution}

\begin{abstract}
We study the problem of unsupervised 3D semantic segmentation on raw point clouds without needing human labels in training. Existing methods usually formulate this problem into learning per-point local features followed by a simple grouping strategy, lacking the ability to discover additional and possibly richer semantic priors beyond local features. In this paper, we introduce \textbf{\nickname{}} to learn 3D semantics from both local and global point features. The key to our approach is to discover 3D semantic information by grouping superpoints according to their global patterns in the frequency domain, thus generating highly accurate semantic pseudo-labels for training a segmentation network. 
Extensive experiments on two indoor and an outdoor datasets show that our 
\nickname{} surpasses all existing unsupervised methods by large margins, achieving the state-of-the-art performance for unsupervised 3D semantic segmentation. Notably, our investigation into the learned global patterns reveals that they truly represent meaningful 3D semantics in the absence of human labels during training. Our code and data are available at {\url{https://github.com/vLAR-group/LogoSP}}
\end{abstract}
\vspace{-0.5cm}

\section{Introduction}
\input{chaps/01_intro.tex}

\section{Related Works}
\input{chaps/02_liter.tex}

\section{\nickname{}}
\input{chaps/03_meth.tex}

\section{Experiments}\label{sec:exp}
\input{chaps/04_exp.tex}

\section{Conclusion}

\input{chaps/05_sum.tex}

\clearpage
{\small
\bibliographystyle{ieeenat_fullname}
\bibliography{reference}}

\clearpage
\appendix
\input{chaps/06_app.tex}
\end{document}

%% file: chaps/01_intro.tex
Semantic segmentation of 3D point clouds is crucial for a variety of emerging applications such as robot planing, embodied AI, and autonomous driving. To learn per-point 3D semantics, early works focus on developing fully or weakly supervised techniques including 3D point-based \cite{Qi2016,Hu2020,Hu2021} methods and 2D/3D projection-based approaches \cite{Graham2018,Wu2017e}. While achieving remarkable accuracy and efficiency on public datasets, they predominantly depend on large-scale human annotations of raw point clouds for training neural networks. Unfortunately, the process of annotating 3D point clouds is particularly costly and labour-intensive. To mitigate this issue, an increasing number of recent works leverage supervised pretrained 3D, 2D/language features such as CLIP \cite{Radford2021} and SAM \cite{Kirillov2023}, followed by finetuning techniques to learn per-point semantics. Despite yielding promising results, these methods still need tedious human annotations or alignments between modalities. 

Very recently, several works tackle 3D semantic segmentation without needing human labels in training, including the seminal GrowSP \cite{Zhang2023a}, PointDC \cite{Chen2023}, and others \cite{Liu2024,Umam2024,Oinonen2024,Ruoppa2025,Zhang2025}. By creating and then progressively growing 3D superpoints within raw point clouds, GrowSP successfully learns semantically meaningful primitives, ultimately achieving purely unsupervised 3D semantic segmentation. Nevertheless, the segmentation performance is still unsatisfactory. The primary reasons are two-fold. First, during its initialization stage, the per-point or per-superpoint features of GrowSP are often not sufficiently discriminative, as its backbone network is randomly initialized without incorporating semantic priors. Second, the latter strategy of grouping 3D superpoints is too simplistic, thus generating inferior pseudo-labels to train the network. 

With the advancement of self-supervised pretraining techniques in 2D domain, semantic hints emerge from pretrained features such as DINO/v2 \cite{Caron2021,Oquab2024} in the absence of any human label during training. In this context, another unsupervised method PointDC \cite{Chen2023} distills these self-supervised 2D features into 3D point clouds, followed by supervoxel grouping for estimating semantics. However, the supervoxel clustering process is error-prone and the resulting soft pseudo-labels often lack accuracy for training the network, which ultimately leads to inferior segmentation performance on 3D point clouds. 

In this paper, we aim to advance the field of unsupervised 3D semantic segmentation. Building on the strengths of prior works, such as GrowSP \cite{Zhang2023a} and PointDC \cite{Chen2023}, while tackling their limitations, we present a simple yet effective pipeline for achieving highly accurate 3D semantic segmentation without needing human annotations in training. Particularly, our pipeline comprises three major ingredients: 1) a 2D-to-3D distillation module that obtains preliminary but high-quality per-point features from self-supervised pretrained models; 2) a bottom-up superpoint growing module that groups raw 3D points into superpoints based on local feature similarity; and 3) a top-down semantic pseudo-label generation module that aggregates superpoints into semantic elements. By constructing a global superpoint graph, this module groups superpoints based on their global features in the frequency domain after applying Graph Fourier Transform. The core of our approach lies in an effective combination of the latter two ingredients, which enables our pipeline to precisely learn per-point semantics through \textbf{lo}cal-\textbf{g}l\textbf{o}bal grouping of \textbf{s}uper\textbf{p}oints without relying on human labels. We name our method \textbf{\nickname{}} and Figure \ref{fig:opening} shows qualitative results for an indoor 3D scene. Our contributions are:
\begin{itemize}[leftmargin=*]
\setlength{\itemsep}{1pt}
\setlength{\parsep}{2pt}
\setlength{\parskip}{1pt} 
    \item We advance the unsupervised semantic segmentation of 3D scene point clouds, seamlessly incorporating self-supervised pretrained 2D features. 
    \item We present an effective technique for top-down grouping superpoints into semantic pseudo-labels based on superpoint features in the global frequency domain.  
    \item We empirically demonstrate the effectiveness of \nickname{} on various datasets, showing significant improvements over all previous unsupervised methods. 
\end{itemize}

%% file: chaps/02_liter.tex
\phantom{Xx}\textbf{Fully-supervised 3D Semantic Learning}: In the past years, there has been tremendous progress in 3D semantic segmentation of point clouds. Current methods can be categorized into four main types: 1) point-based methods \cite{Qi2017,Li2018f,Liu2019h,Hu2020,Hu2021b,Guo2021,Thomas2019,Thomas2024,Wu2019,Zhao2021a,Wang2018c,Kolodiazhnyi2024,Schult2023}, which primarily follow the pioneering work PointNet \cite{Qi2016} to learn per-point features via shared MLPs; 2) 3D voxel-based approaches \cite{Choy2019,Lei2019,Meng2019,Zhu2021b,Peng2024}, which follow the successful SparseConv \cite{Graham2018} to project raw point clouds into regular grids followed by traditional 3D convolutional neural networks; 3) superpoint-based approaches \mbox{\cite{landrieu2018large, robert2023efficient}} which partition the 3D scene into more compact superpoints and assign segmentation results to each superpoint; and 4) 2D projection-based methods \cite{Kundu2020,Milioto2019,Wu2017e}, which project point clouds onto 2D views and leverage established 2D neural architectures. Additionally, the advancement of self-supervised pretraining techniques developed for 3D domain \cite{Rao2020,Xie2020,Huang2021,Zhang2021,Wang2021,Chen2021,Hou2021,Pang2022,Yin2022}  
further enhances segmentation. While achieving impressive results, they rely heavily on human annotations which are prohibitively expensive.

\textbf{Weakly-supervised 3D Semantic Learning}: To alleviate the burden of human labeling, a line of works has been proposed to learn 3D semantics using sparser or cheaper annotations such as fewer 3D point labels \cite{Hu2021,Shi2022,Liu2022,Liu2021,Unal2022,Wu2022,Zhang2021a} or patch-based labels \cite{Wei2020,Chibane2022,Liu2022a}. By further leveraging self-supervised pretraining techniques for point clouds followed by supervised finetuning, many methods \cite{Hou2021,Zhang2021,Xie2020,Zhang2022} have obtained encouraging results. However, they still require human annotations and the trained models are often not generalizable to novel 3D scenes.   

\textbf{Distilling Supervised 2D Features to 3D}: Recently, an increasing number of large 2D vision and/or language models have been developed including SAM/v2 \cite{Kirillov2023,Ravi2024} and CLIP \cite{Radford2021}. Thanks to large-scale 2D datasets with human labels and/or paired language descriptions, these models can extract high-quality visual features from 2D images. Based on them, a series of works \cite{Zhou2022, Ha2022,Rozenberszki2022,Chen2023a,Peng2023,Ding2023,Liu2023,Xiao2024,Thai2024,Xu2024,Yang2024,Peng2024a,Osep2024,Xu2024a,Guo2024,Li2024,Kim2024,Yin2024} have been developed to project 2D features into 3D point clouds, yielding promising results for close-/open-vocabulary 3D semantic learning. 
However, these methods fundamentally rely on human labels in 2D or language domains, 
making them less appealing for applications. In contrast, our method learns 3D semantics without any human labels. 

\textbf{Unsupervised 2D/3D Semantic Learning}: For unsupervised semantic segmentation on 2D images, a number of works have been proposed to learn latent semantic representations by clustering per-pixel features, including the early DeepCluster \cite{Caron2018}, IIC \cite{Ji2019}, PiCIE \cite{Cho2021}, and many recent variants \cite{Gansbeke2021,Hamilton2022,Ouali2020,Zadaianchuk2023,Seong2024,Kim2024a}. 
Although achieving encouraging performance on 2D datasets, few methods show successful applicability in 3D space due to the domain gap between images and point clouds as demonstrated in \cite{Zhang2023a}. For unsupervised 3D semantic learning, early works \cite{Sauder2019,Sun2020} attempt to learn point semantics by recovering voxel positions or canonicalization, but they are limited to object-level point clouds. Very recently, the seminal GrowSP \cite{Zhang2023a} and PointDC \cite{Chen2023} demonstrated encouraging segmentation results on complex real-world 3D datasets without needing any type of human labels in training. In this paper, our new method surpasses them by large margins via effectively grouping superpoints of raw point clouds.  

\begin{figure*}[h!]
\setlength{\abovecaptionskip}{ 2 pt}
\setlength{\belowcaptionskip}{ -12 pt}
\centering
   \includegraphics[width=1.\linewidth]{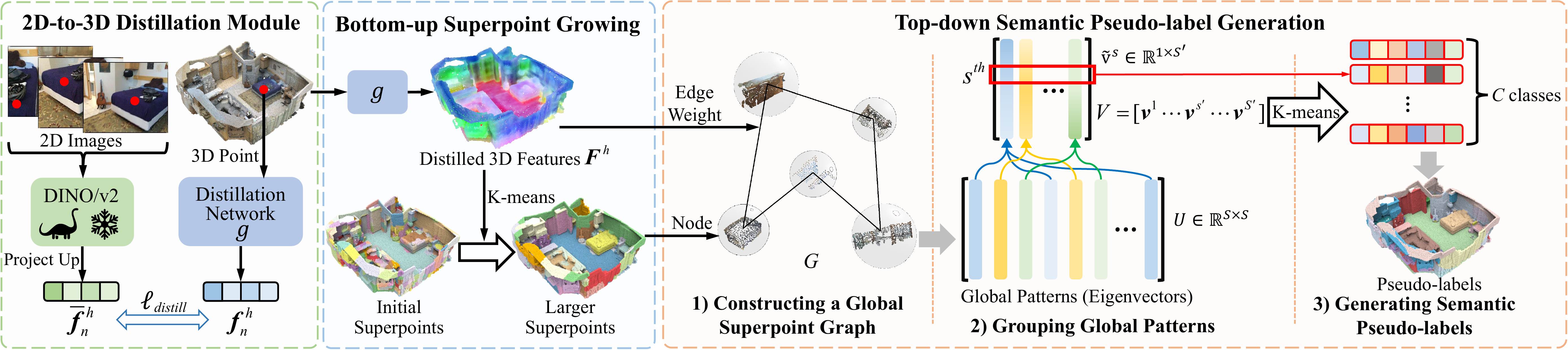}
\caption{The framework of \nickname{}. The leftmost block shows the 2D-to-3D distillation module, the middle block illustrates the bottom-up superpoint growing module, and the rightmost block describes all three steps of our top-down semantic pseudo-label generation module.}
\label{fig:meth_framework}
\end{figure*}

%% file: chaps/03_meth.tex
Given a dataset with $H$ scene point clouds $\{\boldsymbol{P}^1 \cdot\cdot \boldsymbol{P}^h \cdot\cdot \boldsymbol{P}^H\}$, for a specific $h^{th}$ scene point cloud, we have $N$ points, \ie{}, $\boldsymbol{P}^h\in \mathbb{R}^{N\times 6}$, where each point has a location \textit{xyz} with color \textit{rgb}. In the meantime, we also have $K$ RGB/D images $\{\boldsymbol{I}^h_1 \cdot\cdot \boldsymbol{I}^h_k \cdot\cdot \boldsymbol{I}^h_K\}$ associated with the $h^{th}$ point cloud. Such 2D-3D aligned datasets can be easily collected in reality thanks to the wide availability of calibrated 2D/3D scanners on mobile devices or robots. Our goal is to estimate a semantic class belonging to $C$ categories for any $n^{th}$ point $\boldsymbol{p}^h_n$ in $\boldsymbol{P}^h$. In practice, we may safely assume that the total number of semantic classes $C$ is known, as indoor or outdoor scenes often comprise commonly interested classes. 

\subsection{2D-to-3D Distillation Module}\label{sec:2d_to_3d}
As shown in the leftmost block of Figure \ref{fig:meth_framework}, following PointDC \cite{Chen2023}, this module aims to leverage the self-supervised pretrained model in 2D domain to distill semantic priors to 3D. Particularly, given pretrained DINO/v2 \cite{Caron2021,Oquab2024}, for a specific point cloud $\boldsymbol{P}^h$ and its associated 2D images $\{\boldsymbol{I}^h_1 \cdot\cdot \boldsymbol{I}^h_k \cdot\cdot \boldsymbol{I}^h_K\}$, the distillation module has 3 steps:

\textbf{Step 1}: We feed all $K$ images into the pretrained model, obtaining per-pixel features. Following the projection strategy of OpenScene \cite{Peng2023}, these 2D features are projected up to 3D space with the aid of depth images. For every point in $\boldsymbol{P}^h$, if it has multiple feature vectors accumulated, we simply average them out; if it has no feature vector, that point will not be supervised in the subsequent Step 2. 

\textbf{Step 2}: After the above Step 1, for a specific 3D point $\boldsymbol{p}^h_n$, we have its final projected features $\boldsymbol{\bar{f}}^h_n$. Now, we train a SparseConv \cite{Graham2018} from scratch as a distillation network $g$ to regress the projected features for each 3D point. Following OpenScene \cite{Peng2023}, this 3D distillation net takes a single point cloud $\boldsymbol{P}^h$ as input and is supervised by the following loss:
\begin{equation}\label{distill_loss}
\setlength{\abovedisplayskip}{3pt}
\setlength{\belowdisplayskip}{3pt}
\ell_{distill}  = 1 - \cos(\boldsymbol{f}^h_n, \boldsymbol{\bar{f}}^h_n), \quad {\boldsymbol{f}^h_n, \boldsymbol{\bar{f}}^h_n} \in \mathbb{R}^{384}
\end{equation}
where $\boldsymbol{f}^h_n$ is the predicted 3D features from SparseConv. 

\textbf{Step 3}: We obtain a new 3D feature vector for every 3D point in the whole dataset (training split) through the well-trained distillation network $g$. Such 3D features tend to encode more spatial information in addition to semantic priors inherited from 2D images. We reuse the trained network $g$ in subsequent operations.

\subsection{Bottom-up Superpoint Growing}\label{sec:bottomup}
For a specific point cloud $\boldsymbol{P}^h$, we now have its distilled 3D features $\boldsymbol{F}^h\in \mathbb{R}^{N\times 384}$ from net $g$. In this module, we aim to group raw 3D points into larger and larger superpoints that are likely to encode more and more semantic information. As shown in the middle block of Figure \ref{fig:meth_framework}, following GrowSP \cite{Zhang2023a}, we summarize the steps as follows:

\textbf{Step 1}: We create a total number of $M^0$ initial superpoints $\{\boldsymbol{\Tilde{p}}^h_1 \cdots \boldsymbol{\Tilde{p}}^h_{m^0} \cdots \boldsymbol{\Tilde{p}}^h_{M^0}\}$ for a point cloud $\boldsymbol{P}^h$. Note that, \mbox{$M^0$} is actually different across different scenes.

\textbf{Step 2}: We compute mean features for all initial superpoints, denoted as $\{ \boldsymbol{\Tilde{f}}^h_1 \cdot\cdot \boldsymbol{\Tilde{f}}^h_{m^0} \cdot\cdot \boldsymbol{\Tilde{f}}^h_{M^0} \}$: 
\begin{equation}\label{eq:meth_featSP}
\setlength{\abovedisplayskip}{3pt}
\setlength{\belowdisplayskip}{3pt}
\boldsymbol{\Tilde{f}}^h_{m^0} = \frac{1}{Q}\sum_{q=1}^Q \boldsymbol{f}^h_q, \quad \boldsymbol{f}^h_q\in \mathbb{R}^{384}
\end{equation}
where $Q$ is the total number of 3D points within an initial superpoint $\boldsymbol{\Tilde{p}}^h_{m^0}$, and $\boldsymbol{f}^h_q$ is the feature vector retrieved from $\boldsymbol{F}^h$ for the $q^{th}$ 3D point of the superpoint. 

\textbf{Step 3}: With these initial superpoint features, we use K-means to group the $M^0$ vectors into $M^1$ clusters, where $M^1 < M^0$. Each cluster becomes a new and larger superpoint. All new superpoints are:
\begin{equation}
\setlength{\abovedisplayskip}{3pt}
\setlength{\belowdisplayskip}{3pt}
\{\boldsymbol{\Tilde{p}}^h_{1} \cdots \boldsymbol{\Tilde{p}}^h_{m^1} \cdots \boldsymbol{\Tilde{p}}^h_{M^1}\} \xleftarrow{\text{K-means}} \{ \boldsymbol{\Tilde{f}}^h_1 \cdots \boldsymbol{\Tilde{f}}^h_{m^0} \cdots \boldsymbol{\Tilde{f}}^h_{M^0} \}
\end{equation}

With the above three steps, we will have the following larger superpoints in Eq \ref{eq:all_sp} for the whole dataset, and then they are all fed into our third component elaborated in Section \ref{sec:topdown}, generating semantic pseudo-labels to continue training the network $g$ (or a randomly initialized SparseConv) for a predefined $E$ epochs, \ie{}, one round.   
\begin{equation}\label{eq:all_sp}
\setlength{\abovedisplayskip}{3pt}
\setlength{\belowdisplayskip}{3pt}
\Big\{
\{\boldsymbol{\Tilde{p}}^1_{1} \cdots \boldsymbol{\Tilde{p}}^1_{m^1} \cdots \boldsymbol{\Tilde{p}}^1_{M^1}\} \cdots
\{\boldsymbol{\Tilde{p}}^H_{1} \cdots \boldsymbol{\Tilde{p}}^H_{m^1} \cdots \boldsymbol{\Tilde{p}}^H_{M^1}\} 
\Big\}
\end{equation}

After one round, we will calculate the next level of larger superpoints by repeating \textbf{Steps 2/3}. Given $T$ rounds of growth, the number of superpoints for an input point cloud will be decreased from $M^1 \rightarrow M^2$ to a small value $M^T$. For simplicity, we follow GrowSP \mbox{\cite{Zhang2023a}} to assign the same value for $M^t$ across all 3D scenes, except for $M^0$.

\subsection{Top-down Semantic Pseudo-label Generation}\label{sec:topdown}

Having the full set of superpoints of all training scene point clouds, this module is designed to group all superpoints into semantic pseudo-labels, such that we can train a classification network for predicting per-point semantics. To achieve this goal, a na\"ive solution is using K-means to group all superpoints into semantic primitives by comparing the superpoint feature similarity, which is adopted by GrowSP \cite{Zhang2023a}. However, such a simple clustering strategy, fundamentally, can only group per-point local features originally distilled from 2D domain, lacking the ability to capture additional and possibly richer semantic priors beyond local features.

\begin{figure}[h]\vspace{-0.2cm}
\setlength{\abovecaptionskip}{ 2 pt}
\setlength{\belowcaptionskip}{ -6 pt}
\centering
   \includegraphics[width=0.85\linewidth]{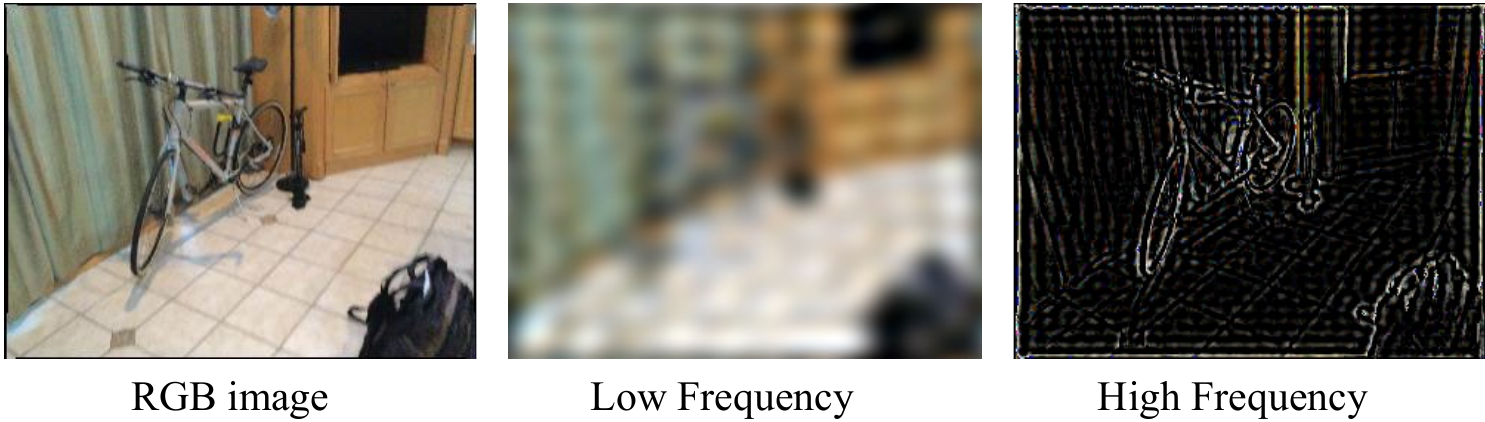}
\caption{An example of high and low frequency features in 2D.}
\label{fig:meth_motivation}
\end{figure}

A complex 3D scene is often composed by many flat planes, curved surfaces, straight edges, and sharp corners, which are crucial to describe 3D semantics. Intuitively, the differences between such features can be well-described in the frequency domain once 3D points are treated as signals. As illustrated in 2D domain in Figure \ref{fig:meth_motivation}, after an image is converted to frequency domain by 2D FFT, a low frequency component tends to indicate the general image background shape, whereas a higher frequency component likely represents shapes of minor objects or details.
\vspace{-0.1cm}

With this insight, we aim to group superpoints in frequency domain which tends to capture semantic priors from a top-down perspective. Since 3D point clouds are irregular and orderless, directly applying 3D FFT is challenging and extremely computationally heavy \cite{hu2022exploring}. In this regard, we turn to convert 3D point clouds into a global graph, and then apply Graph Fourier Transform (GFT) \cite{sandryhaila2013}. As also described in the rightmost block of Figure \ref{fig:meth_framework}, this module involves the following three key steps. \\     

\textbf{1) Constructing a Global Superpoint Graph}

At the very beginning, the whole dataset has $H$ point clouds and each point cloud has $M^0$ superpoints, resulting in a total number of $H*M^0$ superpoints, denoted as $S = H*M^0$. All superpoints have their own features, denoted as $\boldsymbol{\Tilde{F}} \in \mathbb{R}^{S\times 384}$. Then, we build a global superpoint graph by treating each superpoint as a node, and the edge weight $\boldsymbol{a}_{ij}$ between any $i^{th}$ and $j^{th}$ nodes is defined as follows:
\begin{equation}\label{eq:sp_graph_edge}
\setlength{\abovedisplayskip}{3pt}
\setlength{\belowdisplayskip}{3pt}
\boldsymbol{a}_{ij} = \exp\left( -\|\boldsymbol{\Tilde{f}}^i - \boldsymbol{\Tilde{f}}^j\|_2 \right)
\end{equation}
where $\boldsymbol{\Tilde{f}}^i$ and $\boldsymbol{\Tilde{f}}^j$ are the $i^{th}$ and $j^{th}$ superpoint/node features extracted from $\boldsymbol{\Tilde{F}}$ respectively. This gives us a global graph $G$ with $S$ nodes and an adjacency matrix $A \in \mathbb{R}^{S\times S}$ composed by all pairs of $\boldsymbol{a}_{ij}$. 

For the global graph $G$, we compute its normalized Laplacian matrix $L$ as follows: 
\begin{equation}\label{eq:sp_graph_laplacian}
\setlength{\abovedisplayskip}{3pt}
\setlength{\belowdisplayskip}{3pt}
L = D^{-1/2}\cdot(D - A)\cdot D^{-1/2}, \quad L\in \mathbb{R}^{S\times S}
\end{equation}
where $D$ is the diagonal matrix whose entries are the row-wise sum of the adjacency matrix $A$. Following GFT \cite{sandryhaila2013}, the normalized Laplacian matrix $L$ is decomposed into eigenvalues $\Lambda$ and eigenvectors $U$ as follows:
\begin{align}\label{eq:sp_graph_eigen}
\setlength{\abovedisplayskip}{3pt}
\setlength{\belowdisplayskip}{3pt}
L &= U\cdot\Lambda\cdot U^T, \quad \Lambda, U \in \mathbb{R}^{S\times S} \\ \nonumber
U &= [\boldsymbol{u}^1 \cdots \boldsymbol{u}^s \cdots \boldsymbol{u}^S], \quad \boldsymbol{u}^s \in \mathbb{R}^{S\times 1}
\end{align}
where each eigenvector in $U$ represents a basis function or a specific frequency component, which can be naturally regarded as a specific \textit{global pattern} jointly defined by all $S$ superpoints. Clearly, the total $S$ global patterns, \ie{}, the $S$ eigenvectors in $U$, encompass rich semantic priors. \\ 

\textbf{2) Grouping Global Patterns}

Given the total $S$ global patterns (eigenvectors), not all of them are uniquely meaningful in terms of 3D semantics. For example, for the same class \textit{wall} superpoints, there could be multiple global patterns decomposed, where one pattern represents \textit{wall} along $x$-axis and anther pattern $y$-axis. Basically, these global patterns are not discriminative enough in semantics. To this end, we aim to group these patterns.

\textbf{First}, we transform all superpoint features $\boldsymbol{\Tilde{F}}$ to frequency domain via the total global patterns as follows:
\begin{align}\label{eq:sp_graph_feq}
\setlength{\abovedisplayskip}{3pt}
\setlength{\belowdisplayskip}{3pt}
\boldsymbol{\Tilde{F}}_{feq} &= U^T\cdot \boldsymbol{\Tilde{F}}, \quad \boldsymbol{\Tilde{F}}_{feq} \in \mathbb{R}^{S\times 384}\\ \nonumber
\boldsymbol{\Tilde{F}}_{feq} &= [\boldsymbol{f}^1_{feq}\cdots \boldsymbol{f}^s_{feq} \cdots \boldsymbol{f}^S_{feq}]^T, \quad \boldsymbol{f}^s_{feq} \in \mathbb{R}^{1\times 384} 
\end{align}
Mathematically, the $s^{th}$ row vector $\boldsymbol{f}^s_{feq}$ represents the frequency domain features projected from all superpoints' spatial features $\boldsymbol{\Tilde{F}}$, corresponding to the $s^{th}$ global pattern $\boldsymbol{u}^s$. 

\textbf{Second}, we directly group all frequency domain vectors $\{\boldsymbol{f}^1_{feq}\cdot\cdot \boldsymbol{f}^s_{feq} \cdot\cdot \boldsymbol{f}^S_{feq}\}$ into $S'$ clusters using K-means. Then, all the corresponding $S$ global patterns $\{\boldsymbol{u}^1 \cdots \boldsymbol{u}^s \cdots \boldsymbol{u}^S \}$ are assigned to the newly obtained $S'$ clusters.

\textbf{Third}, for each of the new cluster, we merge its assigned global patterns by averaging them out to get a new cluster pattern, \ie{}, a new global pattern as follows:
\begin{equation}\label{eq:sp_group_global}
\setlength{\abovedisplayskip}{3pt}
\setlength{\belowdisplayskip}{3pt}
\boldsymbol{v}^{s'} = \frac{1}{W}\sum_{w=1}^W \boldsymbol{u}^w, \quad \boldsymbol{v}^{s'}\in \mathbb{R}^{S\times 1}
\end{equation}
where $W$ is the total number of global patterns within the $s'$-th cluster. Overall, the original $S$ global patterns are grouped into $S'$ new patterns:
\begin{equation}\label{eq:sp_group_global_v}
\setlength{\abovedisplayskip}{3pt}
\setlength{\belowdisplayskip}{3pt}
\{\boldsymbol{v}^1 \cdot\cdot \boldsymbol{v}^{s'} \cdot\cdot \boldsymbol{v}^{S'}\} \xleftarrow{\text{K-means on $\boldsymbol{\Tilde{F}}_{feq}$}} \{ \boldsymbol{u}^1 \cdot\cdot \boldsymbol{u}^s \cdot\cdot \boldsymbol{u}^S \}
\end{equation}

Our newly grouped global patterns, denoted as $V = [\boldsymbol{v}^1 \cdot\cdot \boldsymbol{v}^{s'} \cdot\cdot \boldsymbol{v}^{S'}]$, tend to have more meaningful semantic priors, fundamentally because the original patterns, which exhibit similar features in frequency domain, have been properly grouped in above steps.

\textbf{3) Generating Semantic Pseudo-labels}

Recall that, from the total $S$ superpoints of the whole dataset, we firstly obtain its original $S$ global patterns $U = [\boldsymbol{u}^1 \cdots \boldsymbol{u}^s \cdots \boldsymbol{u}^S]$, and then group them into $S'$ new global patterns $V = [\boldsymbol{v}^1 \cdots \boldsymbol{v}^{s'} \cdots \boldsymbol{v}^{S'}], (V \in \mathbb{R}^{S\times S'})$. Regarding the new patterns $V$, its $s^{th}$ row vector actually represents how the corresponding $s^{th}$ superpoint contributes to the total $S'$ new global patterns which encompass semantic priors. In other words, for a specific $s^{th}$ superpoint $\boldsymbol{\Tilde{p}}^s$, the $s^{th}$ row vector of $V$, denoted as $\boldsymbol{\Tilde{v}}^s \in \mathbb{R}^{1\times S'}$, can be just regarded as a set of new features from the perspective of new global patterns in frequency domain. 

With this insight, we directly group the total $S$ superpoints into $C$ semantic categories by applying K-means on the row vectors of $V$, thus generating semantic pseudo-labels for all superpoints as follows:
\begin{equation}
\setlength{\abovedisplayskip}{3pt}
\setlength{\belowdisplayskip}{3pt}
\textit{$C$ classes} \xleftarrow{\text{K-means on $V$}} \{\boldsymbol{\Tilde{p}}^1 \cdots \boldsymbol{\Tilde{p}}^s \cdots \boldsymbol{\Tilde{p}}^S\}
\end{equation}

\subsection{Training / Testing}
Given a dataset of point clouds and their associated images together with a self-supervised pretrained DINO/v2, the following Algorithm \ref{alg:training} summarizes how to link all our components to train a SparseConv model for predicting per-point semantics without needing human labels. 

\vspace{-.2cm}
\begin{algorithm}[H]
\caption{ {\small Local-global grouping of superpoints for training.}
}
\label{alg:training}
\begin{algorithmic} 
\footnotesize
\State{\textbf{Input:}} Point clouds $\{\boldsymbol{P}^1 \cdots \boldsymbol{P}^h \cdots \boldsymbol{P}^H\}$, images and DINO/v2;
    
\State{\textbf{Output:}} Per-point semantic predictions belonging to $C$ categories;

\State{\textbf{Init:}} Obtain per-point 3D features via 2D distillation (Sec \ref{sec:2d_to_3d});

\State{\phantom{xx}$\bullet$ Stage 1: Obtain initial superpoints (Step 1 of Sec \ref{sec:bottomup}), and group them into $C$ (pseudo) classes (Sec \ref{sec:topdown});}

\State{\phantom{xx}$\bullet$ Stage 2: Assign pseudo labels to all single 3D points, and train a SparseConv model for $E$ epochs using the standard cross-entropy loss;}

\State{\phantom{xx}$\bullet$ Stage 3: Use the new per-point features from trained SparseConv to grow the sizes of superpoints (Steps 2/3 of Sec \ref{sec:bottomup}); }

\State{\phantom{xx}$\bullet$ Stage 4: Group larger sueprpoints into $C$ (pseudo) classes \scriptsize{(Sec \ref{sec:topdown})}; }

\State{\phantom{xx}$\bullet$ Repeat the above Stages 2/3/4 for $R$ rounds;}

\\ \textbf{Return:} A well-trained SparseConv model to group per 3D point into $C$ classes; Note that, $C$ can be freely chosen in training/test. Our training loss exactly follows GrowSP \cite{Zhang2023a}. 
\end{algorithmic}
\end{algorithm}
\vspace{-.5cm}

After the network is well-trained, our evaluation setting exactly follows GrowSP to infer per-point semantics on test split, without needing to construct any superpoints. All implementation details and hyperparameters are in \cref{sec:app_scannet,sec:app_s3dis,sec:app_unscenes}. 

%% file: chaps/04_exp.tex
\textbf{Datasets}: We evaluate our method on two large-scale real-world indoor datasets:  \textbf{1) ScanNet} \cite{Dai2017} consisting of 1201 / 312 / 100 scenes for training, validation and online test respectively, \textbf{2) S3DIS} \cite{Armeni2017} including six large areas of indoor scenes, and a challenging outdoor dataset: \textbf{3) nuScenes} \cite{Caesar2020} encompassing LiDAR scans of outdoor autonomous driving scenes. We choose the latest self-supervised DINOv2 \cite{Oquab2024} as the pretrained model to distill features. 

\textbf{Baselines}: We select the following baselines: \textbf{1) GrowSP} \citep{Zhang2023a}: this is a pioneering work in unsupervised 3D semantic segmentation, but differs from us in two folds. First, it does not incorporate meaningful semantic priors for initial superpoints, but we leverage the self-supervised DINOv2 as a free lunch. Second, we introduce a new global grouping of superpoints in frequency domain to generate high-quality semantic pseudo-labels. \textbf{2) PointDC} \citep{Chen2023}, it utilizes a 2D pretrained model STEGO \cite{Hamilton2022} based on DINO \cite{Caron2021} for distilling features. For a fair comparison, we implement a DINOv2 version of PointDC, named as \textbf{3) PointDC-DINOv2}. \textbf{4) IIC} \citep{Ji2019}. It is adapted from an existing unsupervised 2D method and modified to use the same backbone as ours. \textbf{5) PiCIE} \citep{Cho2021}. It is also adapted from 2D domain with the same backbone as ours. \textbf{6) K-means} which is applied to cluster per-point \textit{xyzrbg}.

\textbf{Metrics}: The standard metrics mean Intersection-over-Union (\textbf{mIoU}), Overall Accuracy (\textbf{OA}), and mean Accuracy (\textbf{mAcc}) across all classes are reported.

\subsection{Evaluation on ScanNet}\label{sec:exp_scannet}
\begin{figure*}[t]
\setlength{\belowcaptionskip}{ -1 pt}
\centering
   \includegraphics[width=1\linewidth]{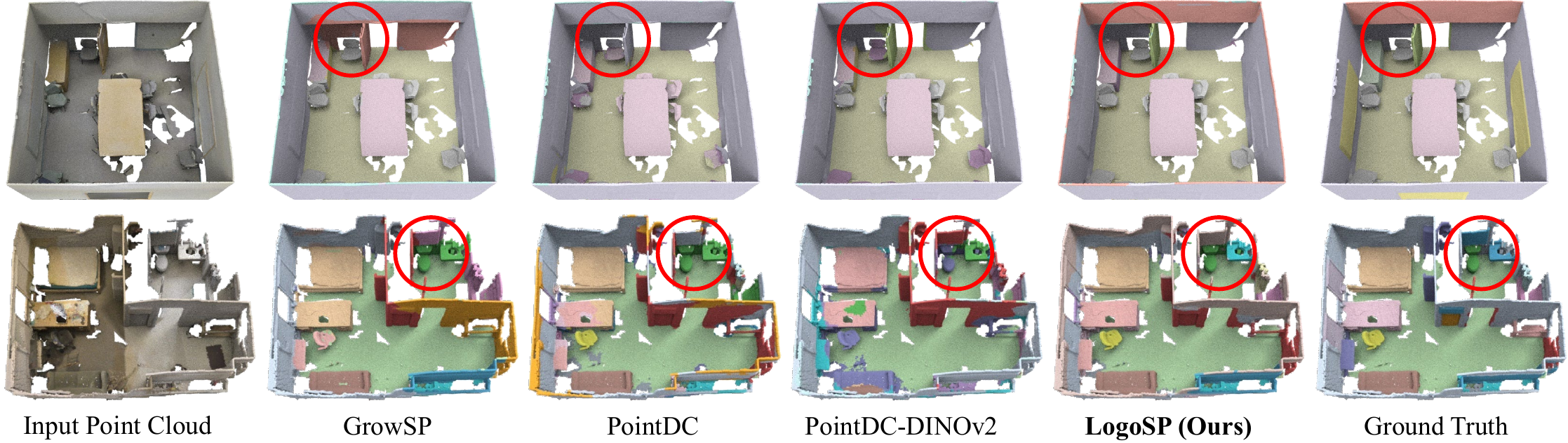}
\caption{Qualitative results on ScanNet dataset (the first row) and S3DIS dataset (the second row). Red circles highlight the differences.}
\label{fig:exp_scannet}
\vspace{-0.3cm}
\end{figure*}

In ScanNet dataset, each point belongs to one of 20 object categories or undefined background. Exactly following GrowSP \cite{Zhang2023a} and PointDC \cite{Chen2023} in testing, we also use Hungarian matching to reorder the predicted labels for calculating scores, reporting results on both the offline validation and the online hidden test sets. The number of global patterns $S'$ is set as 50, and the superpoint growing hyperparameters $M^1$ and $M^T$ are set as 80 and 40 for this dataset, and the number of rounds $T$ as 5. 
Following GrowSP in training, all raw points are fed into the network and undefined points are excluded for loss computation.

\textbf{Results \& Analysis:} As shown in Tables \ref{tab:exp_scannet_val}\&\ref{tab:exp_scannet_online}, our method clearly outperforms all unsupervised baselines by large margins on both val and test sets. PointDC-DINOv2 achieves higher scores than PointDC, but is much weaker than our \nickname{}. This means that our bottom-up and top-down grouping of superpoints can indeed extract meaningful 3D semantics. As shown in Figure \ref{fig:exp_scannet}, we can see that existing baselines tend to group points into major categories, often overlooking minor classes such as \textit{toilet}, whereas our method can effectively identify minor classes primarily because their features in frequency domain can be well captured by our method. More results are in \cref{sec:app_scannet}.

\begin{table}[h]
\centering
\setlength{\belowcaptionskip}{ -6 pt}
\caption{Quantitative results on the \textbf{validation split} of ScanNet dataset \cite{Dai2017}. All 20 categories are evaluated.}
\vspace{-0.3cm}
\label{tab:exp_scannet_val}
\resizebox{0.49\textwidth}{!}{
\begin{tabular}{crccc}
\toprule[1.0pt]
 & & OA(\%) & mAcc(\%) & mIoU(\%) \\
\toprule[1.0pt]
\multirow{6}{*}{\makecell[l]{Unsupervised}}
& K-means  & 10.1 & 10.0 & 3.4 \\
& IIC \citep{Ji2019} & 27.7 & 6.1& 2.9 \\
& PiCIE \citep{Cho2021} & 20.4 & 16.5 & 7.6 \\
& GrowSP \citep{Zhang2023a} & 57.3 & 44.2 & 25.4 \\
& PointDC \citep{Chen2023} & 63.7 & - & 25.7\\
& PointDC-DINOv2 \citep{Chen2023} & 64.7 & 45.0 & 29.6 \\
&\textbf{\nickname{} (Ours)} & \textbf{64.7} &  \textbf{50.8} &  \textbf{35.8} \\
\toprule[1.0pt]
\end{tabular}
}
\vspace{-0.3cm}
\end{table}

\begin{table}[thb]
\centering
\setlength{\abovecaptionskip}{ 2 pt}
\caption{Quantitative results on the \textbf{online hidden split} of ScanNet dataset \cite{Dai2017}. All 20 categories are evaluated.}
\label{tab:exp_scannet_online}
\resizebox{0.33\textwidth}{!}{
\begin{tabular}{crc}
\toprule[1.0pt]
 &  & mIoU(\%) \\
\toprule[1.0pt]
\multirow{3}{*}{\makecell[l]{Supervised}} 
& PointNet++ \cite{Qi2017}  & 33.9 \\
& PointCNN \cite{Li2018f} & 45.8 \\
& SparseConv \cite{Graham2018}  & 72.5 \\
\toprule[1.0pt] 
\multirow{3}{*}{\makecell[l]{Unsupervised}}
& GrowSP \citep{Zhang2023a}  & 26.9 \\ 
& PointDC \citep{Chen2023}  & 22.9 \\ 
&\textbf{\nickname{} (Ours)}  &  \textbf{32.7} \\ 
\toprule[1.0pt]
\end{tabular}
}
 \vspace{-0.7cm}
\end{table}

\subsection{Evaluation on S3DIS}\label{sec:exp_s3dis}
S3DIS comprises 6 large areas with 271 rooms, where all points belong to 13 categories. In training, we follow GowSP to exclude \textit{clutter} points from loss computation and evaluation, as this category exhibits diverse geometric patterns across scenes without consistent semantics. We conduct experiments using standard 6-fold cross validation, training on 5 areas and evaluating on the remaining one. All hyperparameters are the same as used on ScanNet dataset.

\textbf{Results \& Analysis:} As shown in Tables \ref{tab:exp_s3dis_area5}\&\ref{tab:exp_s3dis_6fold}, our method outperforms all baselines on nearly all metrics. 
PointDC-DINOv2 and our method utilize DINOv2 features, but our method significantly surpasses it by 6\% on Area-5 and 5\% on the 6-fold setting on the important mIoU scores. Note that, due to the extremely sparse RGB images collected in this dataset, many 3D points do not correspond to 2D pixels and the distilled 3D features tend to be less discriminative, resulting in our method being just 2\% higher than GrowSP on mIoU. Figure \ref{fig:exp_scannet} compares qualitative results and more results are in \cref{sec:app_s3dis}.

\begin{table}[th] 
\centering
 \setlength{\abovecaptionskip}{ 2 pt}
\caption{Quantitative results of our method and baselines on the Area-5 of S3DIS dataset \cite{Armeni2017}. Only 12 classes are evaluated.}
\label{tab:exp_s3dis_area5}
\resizebox{0.48\textwidth}{!}
{
\begin{tabular}{crccc}
\toprule[1.0pt]
& & OA(\%) & mAcc(\%)& mIoU(\%) \\
\toprule[1.0pt]
\multirow{3}{*}{\makecell[c]{Supervised}} 
& PointNet \cite{Qi2016} & 77.5 & 59.1 & 44.6 \\
& PointNet++ \cite{Qi2017} & 77.5 & 62.6 & 50.1 \\
& SparseConv \cite{Graham2018}& 88.4 & 69.2 & 60.8 \\
\toprule[1.0pt]
\multirow{6}{*}{\makecell[l]{Unsupervised}}
& K-means  & 21.4 & 21.2 & 8.7 \\
& IIC \citep{Ji2019} & 28.5 & 12.5 & 6.4 \\
& PiCIE \citep{Cho2021} & 61.6 & 25.8 & 17.9 \\
& GrowSP \citep{Zhang2023a} & 78.4 & 57.2 & 44.5 \\
& PointDC \citep{Chen2023} & 54.1 & - &22.6 \\
& PointDC-DINOv2 \citep{Chen2023} &75.7 & 48.7 & 40.2\\
&\textbf{\nickname{} (Ours)} &  \textbf{82.8} &  \textbf{55.9} &  \textbf{46.5} \\
\bottomrule[1.0pt]
\end{tabular}
}
\vspace{-0.4cm}
\end{table}

\begin{table}[thb]
\centering
\setlength{\abovecaptionskip}{ 2 pt}
\caption{Quantitative results of 6-fold cross validation on S3DIS dataset \cite{Armeni2017}. Only 12 classes excluding \textit{clutter} are evaluated.}
\label{tab:exp_s3dis_6fold}
\resizebox{0.48\textwidth}{!}{
\begin{tabular}{crccc}
\toprule[1.0pt]
 & & OA(\%) & mAcc(\%) & mIoU(\%) \\
\toprule[1.0pt]
\multirow{3}{*}{\makecell[c]{Supervised}} 
& PointNet \cite{Qi2016} & 75.9 & 67.1 & 49.4 \\
& PointNet++ \cite{Qi2017} & 77.1 & 74.1 & 55.1 \\
& SparseConv \cite{Graham2018}& 89.4 & 78.1 & 69.2 \\
\toprule[1.0pt]
\multirow{6}{*}{\makecell[l]{Unsupervised}}
& K-means  & 20.0 & 21.5 & 8.8 \\
& IIC \citep{Ji2019} & 32.8 & 14.7 & 8.5 \\
& PiCIE \citep{Cho2021} & 46.4 & 28.1 & 17.8 \\
& GrowSP \citep{Zhang2023a} & 76.0 &  \textbf{59.4} & 44.6 \\
& PointDC \citep{Chen2023} & 55.7 & 37.7 & 26.0 \\
& PointDC-DINOv2 \citep{Chen2023} &74.4  &51.5  & 41.3 \\
&\textbf{\nickname{} (Ours)} & \textbf{79.2} &58.0 & \textbf{46.3} \\
\toprule[1.0pt]
\end{tabular}
}
 \vspace{-0.5cm}
\end{table}

\subsection{Evaluation on nuScenes}\label{sec:exp_nuscenes}
We also evaluate our method on a challenging outdoor dataset to demonstrate its generality. The nuScenes dataset contains 16 annotated categories for autonomous driving scenarios, including 28,130 LiDAR scenes aligned with RGB images for training and 6018 scenes for validation. 

\begin{figure*}[t]
\centering
   \includegraphics[width=1\linewidth]{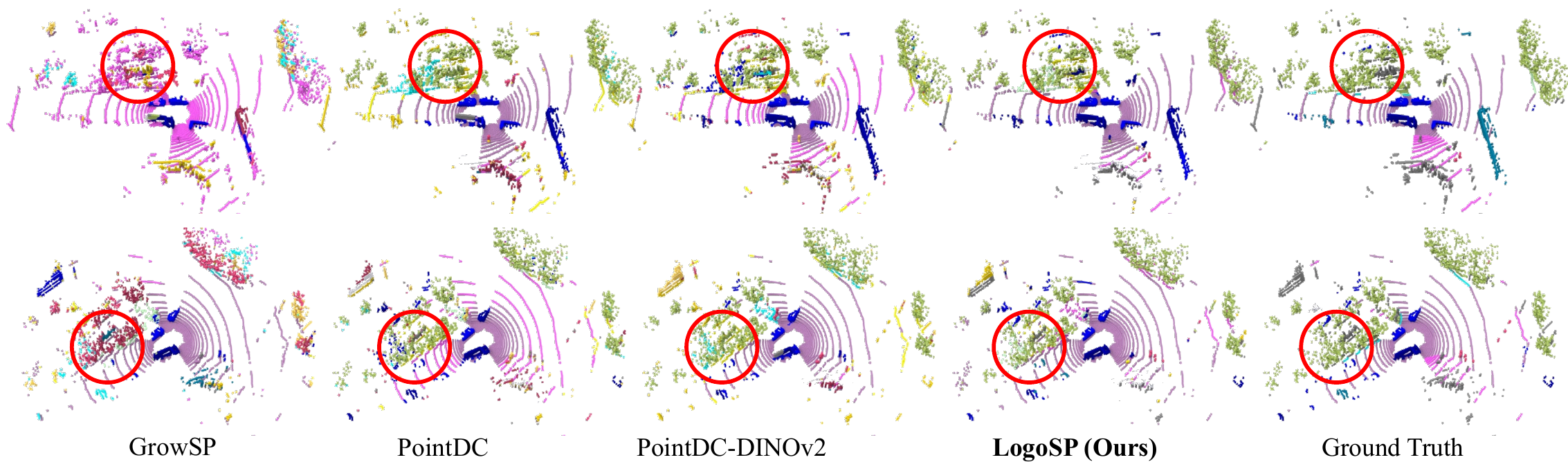}
\caption{Qualitative results on the nuScenes dataset. Red circles highlight the differences.}
\label{fig:exp_nuscenes}
\vspace{-0.3cm}
\end{figure*}

\textbf{Results \& Analysis:} As shown in Table \ref{tab:exp_nuscenes_val}, our method achieves very encouraging results, outperforming all baselines. Notably, our method successfully identifies minor categories such as \textit{person} and \textit{truck}, which are missed by all unsupervised baselines. Figure \ref{fig:exp_nuscenes} shows qualitative results. More results are presented in \cref{sec:app_unscenes}.

\begin{table}[thb]
\centering
\setlength{\belowcaptionskip}{ -6 pt}
\caption{Quantitative results on the \textbf{validation split} of nuScenes dataset \cite{Dai2017}. All 16 categories are evaluated.}
\vspace{-0.3cm}
\label{tab:exp_nuscenes_val}
\resizebox{0.49\textwidth}{!}{
\begin{tabular}{crccc}
\toprule[1.0pt]
 & & OA(\%) & mAcc(\%) & mIoU(\%) \\
\toprule[1.0pt]
\multirow{3}{*}{\makecell[l]{Supervised}} 
&  MinkUNet \citep{choy20194d} & - & - & 73.3 \\
 & Cylinder3D \citep{Zhu2021b} & - & - & 76.1 \\
 & SPVNAS \citep{tang2020searching} & - & - & 77.4 \\
\toprule[1.0pt]
\multirow{4}{*}{\makecell[l]{Unsupervised}}
& GrowSP \citep{Zhang2023a} & 39.2 & 17.5 & 10.2 \\
& PointDC \citep{Chen2023} & \textbf{56.8} &  \textbf{29.4} & 17.7 \\
& PointDC-DINOv2 \citep{Chen2023} & 51.8 & 28.6 & 18.2 \\
&\textbf{\nickname{} (Ours)} &  54.8 & 29.2 &  \textbf{20.1} \\
\toprule[1.0pt]
\end{tabular}
}
\vspace{-0.8cm}
\end{table}

\subsection{Generalization to Unseen Datasets}
To verify that the learned semantic features are generic, we further conduct cross-dataset evaluations as follows:

\noindent \textbf{1) ScanNet to S3DIS}: We directly use the well-trained model on ScanNet in Section \ref{sec:exp_scannet} to extract superpoint features from the ScanNet training set and cluster them into 12 centroids via K-means. These centroids serve as a classifier for the test areas of S3DIS. 

\noindent \textbf{2) S3DIS to ScanNet}: Similarly, we use the well-trained model on S3DIS dataset in Section \ref{sec:exp_s3dis} to obtain 20 centroids from S3DIS training area features, and then evaluate them on the ScanNet validation set.

\textbf{Results \& Analysis:} Table \ref{tab:exp_scannet_2_s3dis} shows results of generalization from ScanNet to S3DIS. We can see that our model trained on the ScanNet dataset significantly outperforms all baselines. Our mIoU scores are over 10\% higher than all methods, primarily because ScanNet contains richer geometry patterns and semantic elements, and our model effectively retains these semantics by grouping global features in the frequency domain. In Table \ref{tab:exp_s3dis_2_scannet}, all methods achieve rather low scores, though our \nickname{} is still the best. The reason is that S3DIS dataset has limited 3D data (just 271 rooms) and sparse 2D images, making the learned semantic priors not generic enough overall.

\begin{table}[thb] \tabcolsep=0.05cm 
\centering
\setlength{\abovecaptionskip}{ 0.5 pt}
\caption{Generalization ability from ScanNet \cite{Dai2017} to the unseen 6 areas of S3DIS \cite{Armeni2017}. The mIoU scores of 12 categories are reported.}
\label{tab:exp_scannet_2_s3dis}
\resizebox{0.49\textwidth}{!}{
\begin{tabular}{rccccccc}
\toprule[1.0pt]
 \textit{test on $\rightarrow$} & Area-1 & Area-2  & Area-3 & Area-4 & Area-5  & Area-6  & mean \\
\toprule[1.0pt]
IIC \citep{Ji2019} &3.7 &3.8 &3.8 &4.0 &3.8 &3.7 &3.8\\
PiCIE \citep{Cho2021} &13.5 &12.7 &13.4 &12.8 &11.3 &13.1 &12.8\\
GrowSP \cite{Zhang2023a} &24.2 &21.9 &26.1 &25.0 &23.7 &27.9 &24.8\\
PointDC \cite{Chen2023} &23.6 &20.9 &24.6 &19.5 &20.1 &29.7 &23.1\\
PointDC-DINOv2 \cite{Chen2023} &33.8 &29.6 &33.8 &31.7 &32.2 &36.9 &33.0\\
\textbf{\nickname{} (Ours)} & \textbf{43.8} & \textbf{37.5} & \textbf{47.0} & \textbf{40.7} & \textbf{44.9} & \textbf{47.9} & \textbf{43.6}\\
\toprule[1.0pt]
\end{tabular}
}
\end{table}

\begin{table}[thb]\tabcolsep=0.02cm 
\centering
\caption{Generalization ability of models trained on different areas of S3DIS \cite{Armeni2017} to the unseen val split of ScanNet \cite{Dai2017}. The mIoU scores of 20 categories are reported.}
\label{tab:exp_s3dis_2_scannet}
\resizebox{0.49\textwidth}{!}{
\begin{tabular}{rcccccccc}
\toprule[1.0pt]
 \multirow{2}{*}{\makecell[c]{\textit{model trained on $\rightarrow$}}} & \multirow{2}{*}{\makecell[c]{Areas\\2/3/4/5/6}} & \multirow{2}{*}{\makecell[c]{Areas\\1/3/4/5/6}} & \multirow{2}{*}{\makecell[c]{Areas\\1/2/4/5/6}} & \multirow{2}{*}{\makecell[c]{Areas\\1/2/3/5/6}} & \multirow{2}{*}{\makecell[c]{Areas\\1/2/3/4/6}}  & \multirow{2}{*}{\makecell[c]{Areas\\1/2/3/4/5}}  & mean\\
  &   &   &   &   &   &   \\
\toprule[1.0pt]
IIC \citep{Ji2019} &3.5 &3.4 &3.7 &3.5 &3.5 &3.6 &3.5\\
PiCIE \citep{Cho2021} &5.6 &5.1 &5.0 &5.9 &6.0 &5.5 &5.5\\
GrowSP \cite{Zhang2023a} &16.9 & \textbf{17.8} &16.4 &16.1 & \textbf{17.1} &15.3 &16.6\\
PointDC \cite{Chen2023} &10.3 &10.1 &10.4 &8.4 &10.0 &9.9 &9.9\\
PointDC-DINOv2 \cite{Chen2023} &16.7 &15.4 &16.4 & \textbf{17.6} &14.3 &14.6 &15.8\\
\textbf{\nickname{} (Ours)} & \textbf{16.9} &16.9 & \textbf{16.8} &16.8 &16.7 & \textbf{17.0} & \textbf{16.9}\\
\toprule[1.0pt]
\end{tabular}
}
\vspace{-0.5cm}
\end{table}

\subsection{Analysis of Global Patterns}
\begin{figure*}[h!]
\setlength{\abovecaptionskip}{ 2 pt}
\setlength{\belowcaptionskip}{ -10 pt}
\centering
   \includegraphics[width=1\linewidth]{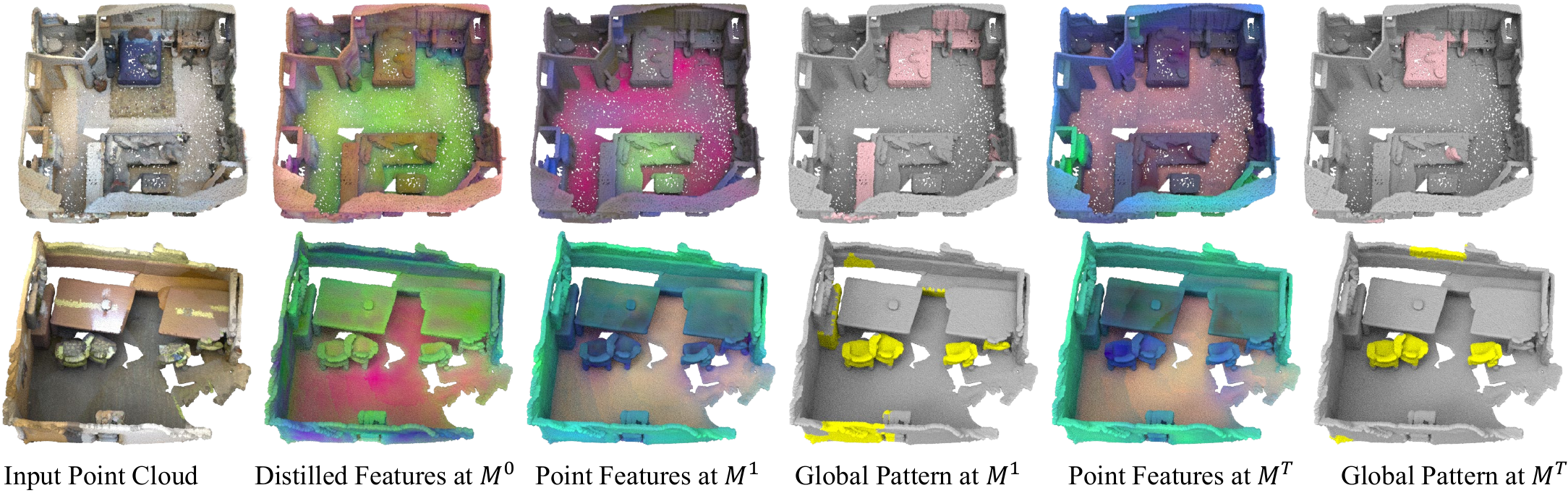}
\caption{Examples of a salient global pattern in each scene  after grouping superpoints in frequency domain, and per-point local features.}
\label{fig:exp_eig}
\end{figure*}

By constructing a global superpoint graph $G$, we first obtain the total $S$ global patterns (eigenvectors) in $U$ via Equation \ref{eq:sp_graph_eigen}. Then we group these global patterns into a new set of $S'$ global patterns in $V$ in Equation \ref{eq:sp_group_global_v}. These new global patterns tend to be much more semantically meaningful. To analyze what these new global patterns really learn, we conduct the following steps to visualize the patterns. 

After training the backbone network on ScanNet dataset for a certain number of epochs, for each scene point cloud, we obtain its $M^t$ superpoints, and we have a total number of $M^t*1201$ superpoints for the whole dataset. Then, we compute the set of $S'$ global patterns $V$ corresponding to the total superpoints. Third, we randomly pick up one of the $S'$ global patterns: \textit{i.e.}, a single $M^t*1201$ dimensional vector where each of the 1201 point clouds contributes an $M^t$ dimensional sub-vector. After that, we randomly select one scene point cloud and its corresponding $M^t$ dimensional sub-vector for analysis. At last, we visualize the relatively high values (salient global patterns in frequency domain) in the $M^t$ dimensional sub-vector, by assigning a bright color to the corresponding superpoints. As shown in Figure \ref{fig:exp_eig}, the 4$^{th}$ and 6$^{th}$ columns illustrate the salient global patterns at $M^1$ and $M^T$ respectively. For a comparison, we also visualize the per-point local features in 3D space, \textit{i.e.}, a 384-dimensional vector, directly predicted from the backbone network. Clearly, we can see that, the global features in frequency domain tend to embed more and more meaningful semantics over the growth of superpoints during training.

\subsection{Ablation Study}
To evaluate the effectiveness of each component of our method and the choices of hyperparameters, we conduct extensive ablation experiments on the ScanNet validation set.

\textbf{(1) Removing 2D-3D distillation}: Instead of training a model to distill 2D features to 3D, we simply copy DINOv2 pixel features to 3D points to grow superpoints and generate pseudo labels, and train a SparseConv from scratch. 

\textbf{(2) $\sim$ (3) Distilling from other self-supervised pretained models}: We also use DINO and STEGO models to distill 3D features in addition to the latest DINOv2.

\textbf{(4) Removing bottom-up superpoint growing}: We evaluate the effectiveness of superpoint growing, as it tends to cover larger areas to enrich the semantics.

\textbf{(5) $\sim$ (8) Different choices of $M^T$}: $M^T$ is the predefined number of superpoints at the end of growing. A smaller $M^T$ indicates aggressive growing with larger superpoints in the end. We set $M^T$ as 40 in main experiments.

\textbf{(9) $\sim$ (11) Different choices of $M^1$}: $M^1$ controls the number of superpoints at the beginning of growing. A greater $M^1$ tends to be safer, while a smaller $M^1$ may incur errors at early steps. We set $M^1$ as 80 in main experiments.

\textbf{(12) Removing semantic pseudo-label generation}: We completely remove our third module top-down semantic pseudo-label generation. Instead, we use K-means to directly group per-point local features. Basically, this variant is GrowSP with distilled features from DINOv2.  

\textbf{(13) $\sim$ (17) Different choices of $S'$}: It controls how aggressive we group global patterns into new ones. Since global patterns in $U$ may be redundant, clustering them could yield more discriminative representations. We set $S'$ as 50 in main experiments.

\textbf{Analysis}: From Table \ref{tab:ablative}, we can see that: 1) The distilled features are crucial to the pipeline as they provide high-quality semantic priors. 2) Superpoint growing is indeed helpful as well and it is relatively robust to its hyperparameters. 3) Without our new top-down grouping of global features in frequency domain, the performance clearly drops, demonstrating the effectiveness of our third component.  

\begin{table}
\centering
\caption{The mIoU scores of all ablated networks on the validation set of ScanNet based on our full \nickname{}.}
\label{tab:ablative}
\resizebox{0.4\textwidth}{!}{
\begin{tabular}{lc}
\toprule[1.0pt]
 & mIoU(\%) \\
\toprule[1.0pt]
(1) Removing distillation &26.8 \\
(2) Distilling DINO features & 25.1\\
(3) Distilling STEGO features &25.8 \\
\toprule[1.0pt]
(4) Removing superpoint growing &29.3 \\
(6) $M^T$ = 50  &34.7  \\
 \textbf{(7) $M^T$ = 40}  & \textbf{35.8}  \\
(8) $M^T$ = 30  &32.4  \\
(9) $M^1$ = 100 &35.7 \\
 \textbf{(10) $M^1$ = 80}  & \textbf{35.8} \\
(11) $M^1$ = 60  &32.1 \\
\toprule[1.0pt]
(12) Removing semantic pseudo-label generation &31.4 \\
(13) $S'$ = 30 &32.4 \\
(14) $S'$ = 40 &32.6 \\
 \textbf{(15) $S'$ = 50} & \textbf{35.8} \\
(16) $S'$ = 60 & 35.6\\
(17) $S'$ = 70 &34.3 \\
\textbf{(18) The full framework (\nickname{})} &\textbf{35.8} \\
\toprule[1.0pt]
\end{tabular}
}
 \vspace{-0.4cm}
\end{table}

%% file: chaps/05_sum.tex
Our proposed method learns per-point semantics from complex 3D scene point clouds without needing human labels in training. By leveraging the high-quality semantic priors distilled from self-supervised 2D features to 3D, we introduce a new global grouping strategy on superpoints to obtain highly accurate semantic pseudo-labels in the frequency domain. Extensive experiments on multiple benchmarks demonstrate the state-of-the-art performance of our method for 3D semantic segmentation. Nevertheless, our performance on the outdoor dataset is mixed, suggesting that DINOv2 features struggle to generalize to self-driving scenes. It would be interesting to extend our method for unsupervised semantic learning on sequential 3D point clouds.

\phantom{xx}

\noindent\textbf{Acknowledgment:} This work was supported in part by National Natural Science Foundation of China under Grant 62271431, in part by Research Grants Council of Hong Kong under Grants 25207822 \& 15225522.

%% file: chaps/06_app.tex
{\large{\noindent\textbf{Appendix}}}

\section{Initial Superpoint Generation}
\begin{figure}[h!]
\centering
   \includegraphics[width=0.87\linewidth]{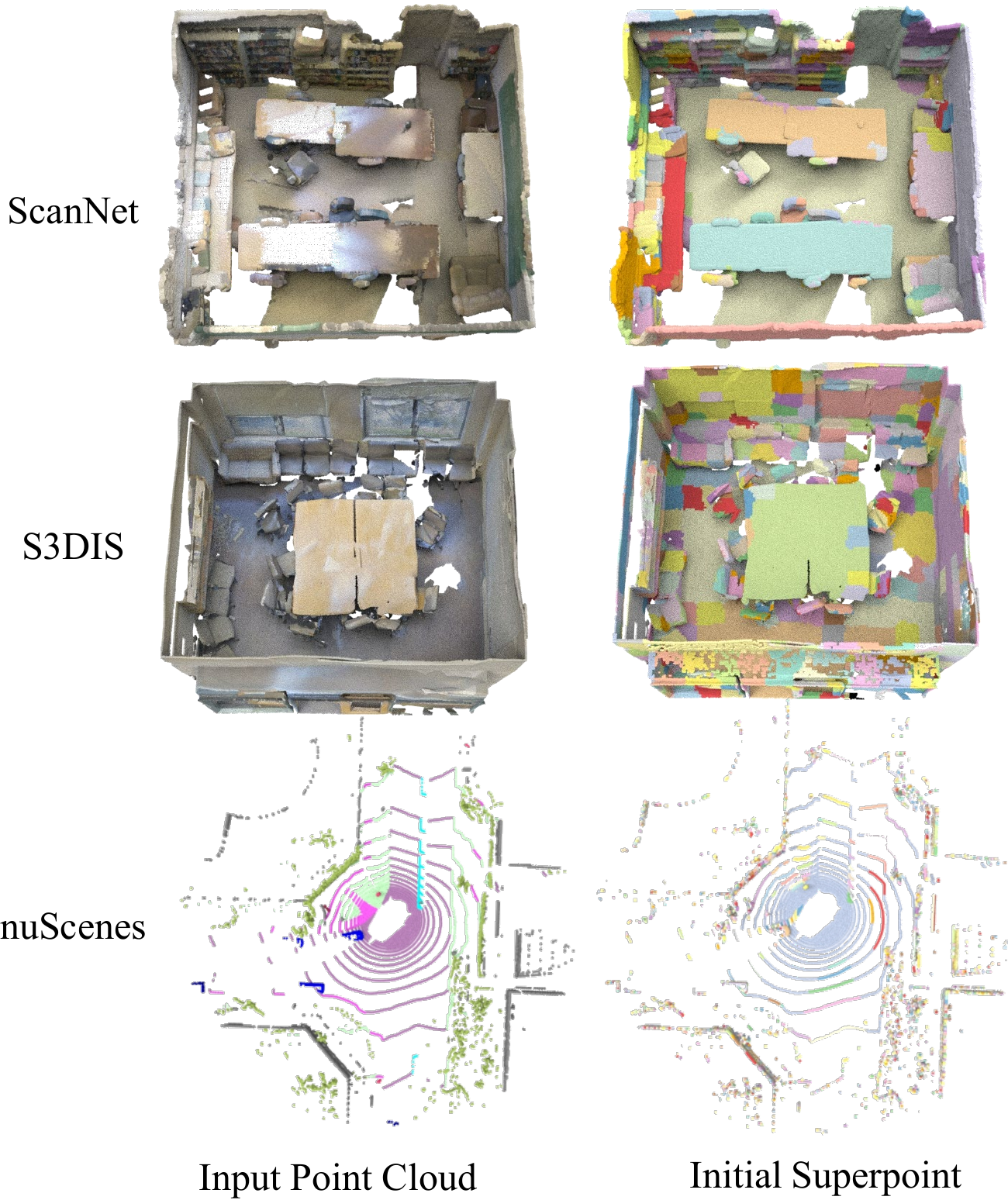}
\caption{Examples of initial superpoints.}
\label{fig:app_init_sp}
\end{figure}

In this section, we describe the detailed steps to generate initial superpoints on three datasets.
For the two indoor datasets ScanNet and S3DIS, we adhere to the steps outlined in GrowSP \cite{Zhang2023a}, utilizing Voxel Cloud Connectivity Segmentation (VCCS) \cite{papon2013voxel} and Region Growing \cite{adams1994seeded} while maintaining the same parameters used in GrowSP. Figure \ref{fig:app_init_sp} illustrates examples of ScanNet and S3DIS datasets.

For the outdoor dataset nuScenes, we adapt from GrowSP by employing RANSAC and Euclidean Clustering to generate superpoints. Initially, RANSAC is applied to identify a large planar surface, designated as the ground. Subsequently, the remaining 3D points are partitioned into clusters using Euclidean Clustering. In RANSAC, points within 0.2 meters of the fitted plane are classified as plane points and aggregated into a single superpoint. For the remaining points, those with an Euclidean distance of less than 0.2 meters are grouped into a single superpoint. Qualitative examples are shown in Figure \ref{fig:app_init_sp}.

\section{Additional Ablation Study}
We conduct additional ablation studies to examine the impact of the semantic category number, distillation process, and segmentation component. 

\textbf{(1) K-means on distilled features}: The distilled point features appear to be aware of semantics, thus we apply K-means on the distilled point features of ScanNet val set to cluster into 20 classes.

\textbf{(2) GrowSP with distillation}: To demonstrate the effectiveness of our global grouping strategy, we apply the well-trained distillation model to GrowSP and complete its subsequent training.

\textbf{(3) $\sim$ (7)}: In our experiments, $C$ is set as 20 for ScanNet, but it can be freely chosen in training. To verify it, we adopt different values: $\{10, 15, 20, 30, 50\}$. 

\textbf{Analysis}: From Table \ref{tab:app_ablative}, we can see that: 1) A simple K-means is not sufficient to discover semantics from the distilled features, so the top-down semantic pseudo-label generation is necessary. 2) GrowSP can be also benefited from the distillation, but the gap between 27.4 and 35.8 indicates our global patterns in the frequency domain is more aligned with semantics. 3) $C$ can be flexible, though too large or too small is not preferred. 

\begin{table}
\centering
\caption{The mIoU scores of all ablated networks on the validation set of ScanNet based on our full \nickname{}.}
\label{tab:app_ablative}
\resizebox{0.4\textwidth}{!}{
\begin{tabular}{lc}
\toprule[1.0pt]
 & mIoU(\%) \\
\toprule[1.0pt]
(1) K-means on distilled features &18.2 \\
(2) GrowSP with distillation & 27.4\\
\toprule[1.0pt]
(3) $C$ = 10 &32.8 \\
(4) $C$ = 15 &33.7 \\
 \textbf{(5) $C$ = 20} & \textbf{35.8} \\
(6) $C$ = 30 & 34.7\\
(7) $C$ = 50 &32.9 \\
\textbf{(8) The full framework (\nickname{})} &\textbf{35.8} \\
\toprule[1.0pt]
\end{tabular}
}
\end{table}

\section{Impacts of Initial Superpoint Purity}
The initial superpoints are crucial in our \nickname{}. If the initial superpoints extensively cover points from different categories, it can lead to a decline in segmentation performance. To measure this, we define the purity of superpoints as follows: we use ground truth semantic labels to assign a unique label to each superpoint through a voting process. Consequently, all points receive a label generated from this voting. We then compute the mIoU between these voting labels and the ground truth to assess the purity of initial superpoints.

The initial superpoints are constructed by VCCS and Region Growing, the smaller resolutions of VCCS yield purer superpoints, and the higher the mIoU against GT, the purer. As shown in Table \ref{tab:exp_initial_purity}, we obtain robust results even at rather low purity on ScanNet val.

Another interesting observation is the trend of purity over the growing process. From Table \ref{tab:exp_trend}, when $M^i$ is reduced to $M^0\rightarrow$ 80/70/60/50/40,
the superpoint purity (mIoU) drops 76.8 $\rightarrow$ 74.6/73.8/72.6/71.4/69.4 as expected.

\begin{table}[h]
\setlength{\abovecaptionskip}{ 2 pt}
\setlength{\belowcaptionskip}{ -10 pt}
\begin{center}
\caption{Impact of initial superpoint purity on ScanNet val.}
\label{tab:exp_initial_purity}
\resizebox{0.47\textwidth}{!}{
\begin{tabular}{lcccccc}
\toprule[1.0pt]
resolution (m)  &0.1 &0.3  &0.5 &0.7 &0.9 \\
\toprule[1.0pt]
Initial superpoint purity (mIoU) &83.9 &81.7  &76.8  &71.7  &66.1 \\
Segmentation results (mIoU)        &34.7 &35.9  &35.8  &34.8  &31.3 \\
\toprule[1.0pt]
\end{tabular}}
\end{center}
\end{table}\vspace{-0.4cm}

\begin{table}[h]
\setlength{\abovecaptionskip}{ 2 pt}
\setlength{\belowcaptionskip}{ -10 pt}
\begin{center}
\caption{Trend of purity over growing.}
\label{tab:exp_trend}
\resizebox{0.47\textwidth}{!}{
\begin{tabular}{lcccccc}
\toprule[1.0pt]
Number of superpoints ($M^i$) &$M^0$ &80 &70  &60  &50  &40 \\
\toprule[1.0pt]
Superpoints purity (mIoU)        &76.8 &74.6  &73.8  &72.6  &71.4 &69.4 \\
\toprule[1.0pt]
\end{tabular}}
\end{center}
\end{table}\vspace{-0.4cm}

\section{Evaluation on ScanNet}
\label{sec:app_scannet}

\begin{figure*}[t!]
\centering
   \includegraphics[width=1\linewidth]{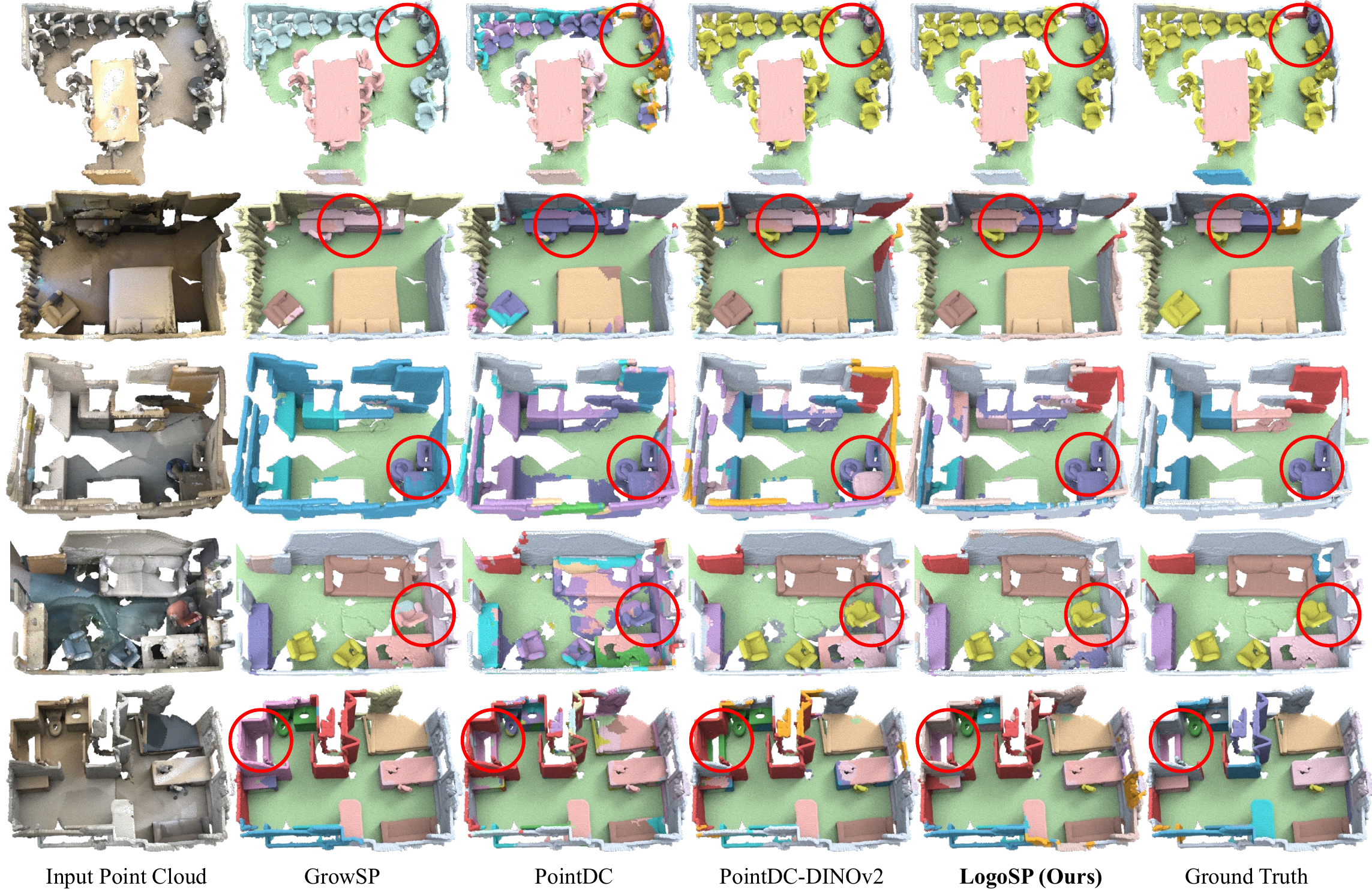}
\caption{Qualitative results of our method and baselines on the validation set of ScanNet dataset.}
\label{fig:app_scannet}
\end{figure*}

When evaluating on the ScanNet dataset, we utilize SparseConv as our backbone which is the same as GrowSP\cite{Zhang2023a}. A voxel size of 5cm is employed to convert point clouds into voxel grids for both distillation and segmentation tasks. For distillation, we select the ViT-S/14 version of DINOv2 as the 2D feature extractor for our method and the baseline PointDC-DINOv2. The distillation process is conducted over 200 epochs with a batch size of 8, a learning rate of 1e-3, and the Adam optimizer. The Poly scheduler is used to progressively decrease the learning rate. For training our segmentation network, we employ cross-entropy as the loss function, maintaining a batch size of 8 and a constant learning rate of 1e-4 with the Adam optimizer over 200 epochs. The superpoints growing parameters $M^1$ and $M^T$ are set as 80 and 40.
The extraction of global patterns and
generation of pseudo labels are conducted every 10 epochs.

The per-category results on both validation and hidden test sets are detailed in Tables \ref{tab:scannet_val_detail}\&\ref{tab:scannet_test_detail}. Our method significantly outperforms all unsupervised baselines, particularly on minor classes such as \textit{books}, \textit{curtain}, and \textit{toilet}. Figure \ref{fig:app_scannet} provides further comparisons with baselines.

\begin{table*}[t!]
\centering
\caption{Per-category quantitative results on the validation split of ScanNet dataset.}
\label{tab:scannet_val_detail}
\resizebox{\textwidth}{!}
{
\begin{tabular}{rccccccccccccccccccccccc}
\toprule[1.0pt]
& OA(\%) & mAcc(\%)& mIoU(\%) & wall. & floor. & cab. & bed. & chair. & sofa. & table & door. & wind. & books. & pic. & counter. & desk. & curtain. & fridge. & shower. & toilet. & sink. & bathtub. & otherf.\\
\toprule[1.0pt]
\multirow{7}{*} 

K-means  &10.1 &10.0 &3.4 &9.0 &9.8 &3.2 &2.9 &5.5 &3.3 &4.3 &3.5 &5.5 &3.3 &\textbf{2.6} &0.8 &2.9 &4.3 &0.8 &0.8 &0.7 &0.3 &0.9 &4.0\\

IIC \citep{Ji2019}  &27.7 &6.1 &2.9 &25.3 &20.5 &0.6 &0.3 &3.7 &0.4 &1.3 &1.3 &1.1 &1.9 &0.2 &0.1 &0.6 &0.3 &0.4 &0 &0 &0 &0.2 &0.5\\

PICIE \citep{Cho2021}  &20.4 &16.5 &7.6 &14.7 &24.5 &6.3 &5.2 &18.0 &8.4 &33.2 &6.7 &4.8 &9.3 &2.1 &0.1 &2.7 &8.0 &1.1 &2.1 &0 &0 &0.5 &5.0\\

GrowSP \citep{Zhang2023a}  &57.3 &44.2 &25.4 &40.7 &89.8 &\textbf{24.0} &47.2 &45.5 &43.0 &39.4 &14.1 &20.0 &53.5 &0.1 &5.4 &13.3 &8.4 &\textbf{2.1} &\textbf{11.3} &20.6 &\textbf{19.4} &0 &9.8 \\

PointDC \citep{Chen2023}  &62.4 &38.8 &26.0 &\textbf{59.1} &\textbf{94.0} &22.0 &43.2 &30.4 &35.9 &38.3 &14.3 &37.4 &44.4 &1.2 &2.4 &2.3 &\textbf{39.4} &2.0 &0 &38.6 &0 &2.2 &12.7 \\
PointDC-DINOv2 \citep{Chen2023}  &64.7 &45.0 &29.6 &57.0 &86.0 &18.5 &60.6 &60.1 &46.2 &46.4 &27.0 &39.0 &54.2 &0.7 &25.0 &\textbf{18.1} &22.7 &0.2 &2.8 &16.8 &0 &0 &10.4\\
\textbf{\nickname{} (Ours)}&\textbf{64.7} &\textbf{50.8} &\textbf{35.8} &46.3 &86.6 &20.7 &\textbf{66.8} &\textbf{63.3} &\textbf{50.9} &\textbf{47.1} &\textbf{33.8} &\textbf{41.6} &\textbf{62.8} &1.0 &\textbf{38.0} &10.5 &28.6 &0.5 &0 &\textbf{46.3} &0 &\textbf{42.3} &\textbf{29.6}\\
\bottomrule[1.0pt]
\end{tabular}
}
\end{table*}

\begin{table*}[t!]
\centering
\caption{Per-category quantitative results on the hidden test split of ScanNet dataset.}
\label{tab:scannet_test_detail}
\resizebox{\textwidth}{!}
{
\begin{tabular}{crccccccccccccccccccccc}
\toprule[1.0pt]
& &mIoU(\%) & wall. & floor. & cab. & bed. & chair. & sofa. & table & door. & wind. & books. & pic. & counter. & desk. & curtain. & fridge. & shower. & toilet. & sink. & bathtub. & otherf.\\
\toprule[1.0pt]
\multirow{3}{*}{\makecell[c]{Supervised} }
& PointNet++ \cite{Qi2017} &33.9 &52.3 &67.7 &25.6 &47.8 &36.0 &34.6 &23.2 &26.1 &25.2 &45.8 &11.7 &25.0 &27.8 &24.7 &18.3 &14.5 &54.8 &36.4 &58.4 &18.3\\
& DGCNN \cite{Wang2018c}  &44.6 &72.3 &93.7 &36.6 &62.3 &65.1 &57.7 &44.5 &33.0 &39.4 &46.3 &12.6 &31.0 &34.9 &38.9 &28.5 &22.4 &62.5 &35.0 &47.4 &27.1\\
& PointCNN \cite{Li2018f} &45.8 &70.9 &94.4 &32.1 &61.1 &71.5 &54.5 &45.6 &31.9 &47.5 &35.6 &16.4 &29.9 &32.8 &37.6 &21.6 &22.9 &75.5 &48.4 &57.7 &28.5\\
& SparseConv \cite{Graham2018}  &72.5 &86.5 &95.5 &72.1 &82.1 &86.9 &82.3 &62.8 &61.4 &68.3 &84.6 &32.5 &53.3 &60.3 &75.4 &71.0 &87.0 &93.4 &72.4 &64.7 &57.2\\
\toprule[1.0pt]
\multirow{1}{*}{\makecell[c]{Unsupervised} } 
&GrowSP \cite{Zhang2023a} & 26.9 & 32.8 & \textbf{89.6} & 15.2 & 62.9 & 55.3 & 38.9 & 32.0 & 14.4 & 23.0 & \textbf{59.9} & 0 & 12.5 & \textbf{11.4} & 6.1 & \textbf{1.2} & \textbf{9.3} & 43.9 & \textbf{14.0} & 0 & 16.5 \\
&\textbf{\nickname{}(Ours)}& \textbf{32.7} & \textbf{41.4} & 87.1 & \textbf{18.1} & \textbf{68.4} & \textbf{56.2} & \textbf{49.9} & \textbf{39.6} & \textbf{30.2} & \textbf{48.7} & 49.2 & \textbf{0.1} & \textbf{29.1} & 7.3 & \textbf{33.4} & 0 & 0 & \textbf{54.3} & 0 & \textbf{21.1} & \textbf{19.3} \\
\bottomrule[1.0pt]
\end{tabular}
}
\end{table*}

\section{Evaluation on S3DIS}
\label{sec:app_s3dis}

\begin{figure*}[t!]
\centering
   \includegraphics[width=1\linewidth]{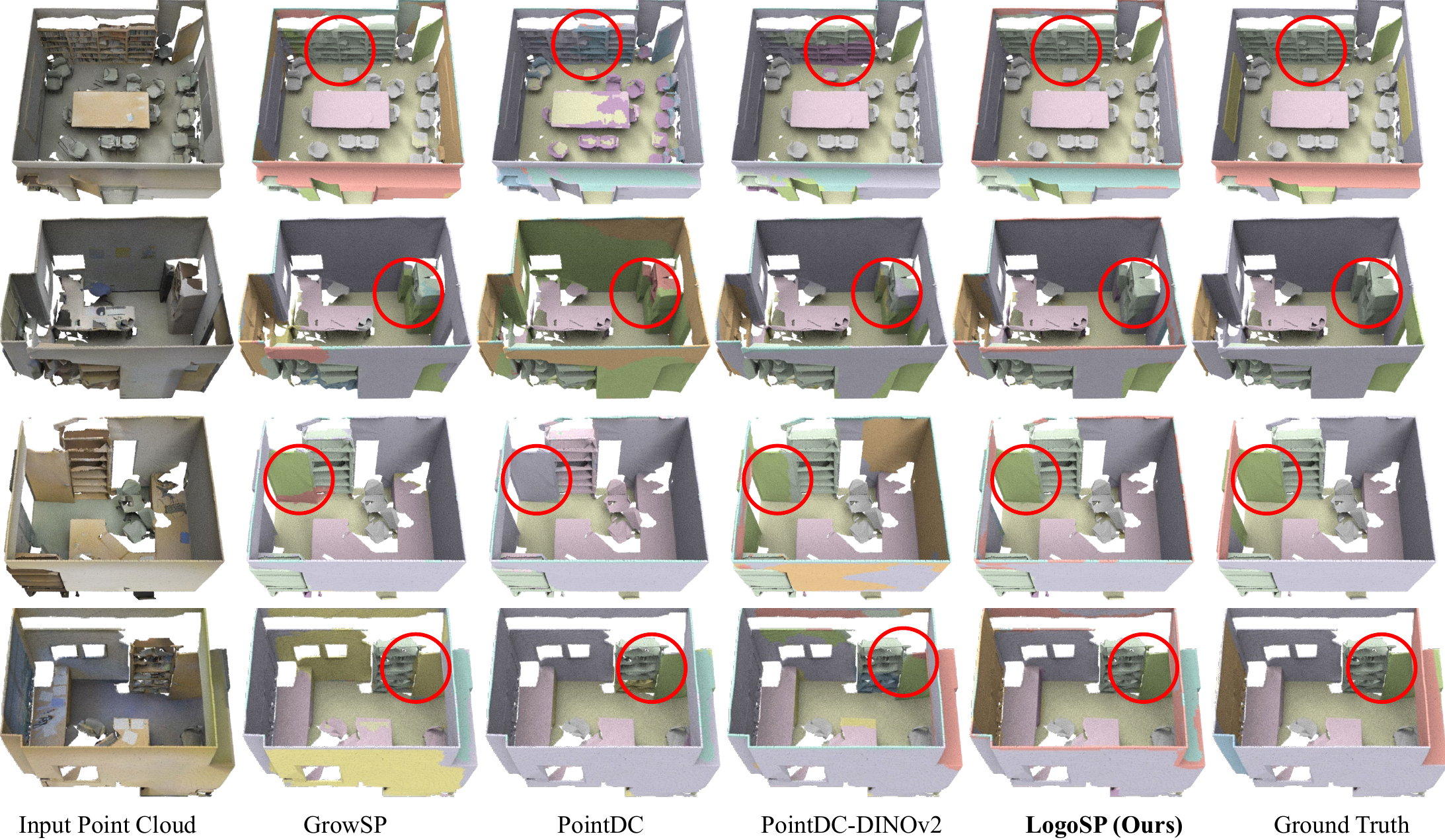}
\caption{Qualitative results of our method and baselines on the S3DIS dataset.}
\label{fig:app_s3dis}
\end{figure*}

Prior to training, we downsample input point clouds by applying grid sampling with a 0.01m grid size. Distillation and segmentation are then performed on the downsampled data. Same as ScanNet, we also choose ViT-S/14 of DINOv2 model in the configuration. For distillation, we choose the learning rate of 1e-3, the Adam optimizer, and the Poly scheduler. When training the segmentation network, we employ a learning rate of 1e-4 over 200 epochs, with the parameter $S'$ for grouping global patterns being 10. $M^1$ and $M^T$ are also set as 80 and 40 respectively.

The per-category results for each area and the 6-fold evaluation are presented in Tables \ref{tab:s3dis_a1} 
to \ref{tab:s3dis_6fold}. Our method demonstrates improvements across all areas. Figure \ref{fig:app_s3dis} shows qualitative results. Since PointDC does not provide detailed results for each category, we reproduce results by training its own models.

\begin{table*}[t!]
\centering
\caption{Quantitative results of our method and baselines on the Area-1 of S3DIS dataset.}
\label{tab:s3dis_a1}
\resizebox{\textwidth}{!}
{
\begin{tabular}{crccccccccccccccc}
\toprule[1.0pt]
& & OA(\%) & mAcc(\%)& mIoU(\%) & ceil. & floor & wall & beam & col. & wind. & door & table & chair & sofa & book. & board. \\
\toprule[1.0pt]
\multirow{3}{*}{\makecell[l]{Supervised} }
& PointNet \cite{Qi2016} & 75.4 & 74.8 & 55.0 & 88.3 & 93.2 & 69.2 & 49.5 & 37.8 & 74.5 & 65.6 & 41.2 & 42.5 & 22.3 & 35.4 & 40.9\\

& PointNet++ \cite{Qi2017} & 76.1 & 77.9 & 58.2 & 90.5 & 94.4 & 65.7 & 38.2 & 31.9 & 61.5 & 66.0 & 45.3 & 60.4 & 41.2 & 45.8 & 57.4\\

& SparseConv \cite{Graham2018} &89.0 & 79.5& 72.5 & 93.6 & 95.6 & 76.1 & 65.9 & 60.9 & 60.0 & 74.2 & 81.9 & 85.4 & 69.2 & 73.4 & 33.5\\
\toprule[1.0pt]
\multirow{8}{*}{\makecell[c]{Unsupervised} } 
& Kmeans  & 20.9 & 24.1 & 10.1 & 15.4 & 17.8 & 10.5 & 16.8 & 1.9 & 16.0 & 12.1 & 9.9 & 8.1 & 0.1 & 6.2 & 6.7\\
& IIC \citep{Ji2019} & 29.2 & 14.3 & 8.0 & 17.0 & 31.4 & 25.6 & 4.3 & 11.1 & 0 & 2.6 & 1.4 & 0.7 & 0 & 0.2 & 1.4\\
& PICIE \citep{Cho2021} & 45.7 & 28.3 & 19.4 & 77.2 & 63.1 & 24.5 & 15.8 & 3.3 & 4.4 & 9.6 & 10.2 & 14.7 & 0 & 9.9 & 0\\
& GrowSP \citep{Zhang2023a} & 72.9 & \textbf{60.4} & 45.6 & \textbf{94.2} & 90.8 & 52.7 & \textbf{36.7} & 19.7 & 33.3 & 35.8 & 66.5 & \textbf{72.6} & 13.1 & 31.2 & \textbf{16.7} \\
& PointDC \citep{Chen2023} & 58.0 & 41.5 & 28.8 & 88.7 & 89.5 & 31.9 & 1.5 & 7.1 & 17.6 & 12.4 & 46.4 & 17.0 & 0 &32.7 & 0\\
& PointDC-DINOv2 \citep{Chen2023} & 73.8 & 55.4 & 44.0 & 85.7 & \textbf{93.7} & 58.9 & 10.1 & \textbf{20.2} & 0 & \textbf{45.0} & \textbf{70.1} & 61.5 & \textbf{44.0} & 39.4 & 0 \\
&\textbf{\nickname{} (Ours)}& \textbf{76.9} & 60.0 & \textbf{48.9} & 89.0 & 93.2 & \textbf{63.2} & 27.5 & 19.2 & \textbf{71.3} & 39.7 &69.7 & 69.2 & 0.7 & \textbf{43.6} & 0 \\
\bottomrule[1.0pt]
\end{tabular}
}
\end{table*}

\begin{table*}[t!]
\centering
\caption{Quantitative results of our method and baselines on the Area-2 of S3DIS dataset.}
\label{tab:s3dis_a2}
\resizebox{\textwidth}{!}
{
\begin{tabular}{crccccccccccccccc}
\toprule[1.0pt]
& & OA(\%) & mAcc(\%)& mIoU(\%) & ceil. & floor & wall & beam & col. & wind. & door & table & chair & sofa & book. & board. \\
\toprule[1.0pt]
\multirow{3}{*}{\makecell[l]{Supervised} }
& PointNet \cite{Qi2016} & 72.5 & 55.5 & 36.6 & 79.2 & 87.4 & 64.9 & 14.5 & 8.2 & 14.8 & 39.6 & 28.8 & 64.0 & 7.8 & 24.4 & 5.1\\

& PointNet++ \cite{Qi2017} & 72.1 &62.3 & 39.9 & 85.8 & 69.6 & 71.2 & 24.9 & 27.5 & 32.5 & 43.6 & 27.4 & 51.3 & 6.0 & 26.8 &12.4 \\

& SparseConv \cite{Graham2018} & 87.9 & 69.5 & 57.3 & 89.5 & 93.8 & 77.0 & 29.1 & 32.5 & 65.5 & 45.7 & 67.9 & 88.8 & 34.9 & 54.5 & 8.2\\
\toprule[1.0pt]
\multirow{8}{*}{\makecell[c]{Unsupervised} } 
& Kmeans  & 17.6 & 16.6 & 6.4 & 16.4 & 15.6 & 11.3 & 3.3 & 0.9 & 0.4 & 6.8 & 3.7 & 11.0 & 1.4 & 4.6 & 1.5\\
& IIC \citep{Ji2019} & 41.6 & 16.8 & 10.6 & 33.0 & 43.7 & 27.6 & 1.7 & 0 & 0 & 5.6 & 0.1 & 13.0 & 0 & 2.8 & 0\\
& PICIE \citep{Cho2021} & 48.3 & 27.2 & 17.4 & 72.4 & 44.2 & 39.6 & 6.2 & 1.7 & 0.5 & 7.7 & 4.1 & 20.1 & 0 & 7.7 & 3.6\\
& GrowSP \citep{Zhang2023a} & \textbf{79.0} & 51.8 & 39.1 & 85.7 & 88.2 & 67.0 & \textbf{12.0} & \textbf{24.8} & 0 & 24.2 & 51.2 & 77.1 & \textbf{4.1} & 24.5 & 0.2\\
& PointDC \citep{Chen2023} & 48.3 & 34.8 & 22.5 & 66.7 & 50.2 & 26.1 & 1.3 & 0.4 & 0 & 15.6 & 29.8 & 56.3 & 0.3 & 17.6 & \textbf{5.6} \\
& PointDC-DINOv2 \citep{Chen2023}& 77.1 & 50.3 & 38.1 & 90.8 & \textbf{92.0} & 57.3 & 10.2 & 0.6 & 34.2 & 20.3 & 46.4 & \textbf{85.6} & 0 & 19.1 & 0.9\\
&\textbf{\nickname{} (Ours)} & 77.0 & \textbf{52.2} & \textbf{39.4} & \textbf{92.3} & 67.7 & \textbf{72.0} & 7.2 & 0.7 & \textbf{44.8} & \textbf{32.4} & \textbf{58.0} & 53.3 & 0.2 & \textbf{44.1} & 0.5\\
\bottomrule[1.0pt]
\end{tabular}
}
\end{table*}

\begin{table*}[t!]
\centering
\caption{Quantitative results of our method and baselines on the Area-3 of S3DIS dataset.}
\label{tab:s3dis_a3}
\resizebox{\textwidth}{!}
{
\begin{tabular}{crccccccccccccccc}
\toprule[1.0pt]
& & OA(\%) & mAcc(\%)& mIoU(\%) & ceil. & floor & wall & beam & col. & wind. & door & table & chair & sofa & book. & board. \\
\toprule[1.0pt]
\multirow{3}{*}{\makecell[l]{Supervised} }
& PointNet \cite{Qi2016} & 78.2 & 74.9 & 57.7 & 90.3 & 96.9 & 66.9 & 55.5 & 15.1 & 60.0 & 67.7 & 51.8 & 54.8 & 27.6 & 56.0 & 50.0\\

& PointNet++ \cite{Qi2017} & 79.8 & 85.9 & 65.8 & 91.4 & 98.0 & 68.5 & 50.1 & 15.2 & 74.8 & 74.7 & 63.2 & 70.1 & 53.6 & 54.0 & 76.5\\

& SparseConv \cite{Graham2018} & 91.3 & 86.8 & 78.6 & 93.1 & 96.2 & 80.4 & 74.7 & 63.3 & 77.2 & 69.5 & 80.1 & 85.5 & 89.5 & 80.1 & 52.5\\
\toprule[1.0pt]
\multirow{8}{*}{\makecell[c]{Unsupervised} } 
& Kmeans  & 21.3 & 22.1 & 9.4 & 20.2 & 20.6 & 13.3 & 5.7 & 1.3 & 2.3 & 14.1 & 6.8 & 6.8 & 3.7 & 9.7 & 8.6\\
& IIC \citep{Ji2019} & 32.1 & 15.4 & 8.4 & 20.5 & 25.5 & 31.4 & 1.0 & 6.9 & 0.2 & 3.2 & 1.6 & 0.3 & 0 & 10.6 & 0\\
& PICIE \citep{Cho2021} & 40.4 & 29.2 & 16.2 & 50.5 & 49.6 & 33.7 & 13.2 & 3.0 & 1.8 & 6.5 & 8.9 & 7.5 & 3.5 & 16.2 & 0.4\\
& GrowSP \citep{Zhang2023a} & 74.2 & \textbf{68.4} & 47.7 & \textbf{92.9} & 91.7 & 48.3 & \textbf{49.3} & 15.8 & 21.1 & 38.7 & 60.6 & \textbf{66.5} & \textbf{28.5} & 59.2 & 0\\
& PointDC \citep{Chen2023} & 56.2 & 40.7& 27.2 & 75.3 & 91.3 & 29.7 & 1.2 &  2.2 & 0 & 11.4 & 37.9 &  20.7 & 9.2 & 38.4 & \textbf{8.9}\\
& PointDC-DINOv2 \citep{Chen2023} & 70.5 &52.4 & 39.3 & 86.2 & 93.4 & 48.8 &0 &14.5 & 38.6 &28.2 &\textbf{68.7} & 55.0 & 0 & 37.7 & 0\\
&\textbf{\nickname{} (Ours)} & \textbf{79.8} & 62.7 & \textbf{48.9} & 90.0 & \textbf{94.3} & \textbf{65.9} & 16.0 & \textbf{18.6} & \textbf{67.8} & \textbf{45.5} & 59.2 & 56.3 & 3.6 & \textbf{69.4} & 0\\
\bottomrule[1.0pt]
\end{tabular}
}
\end{table*}

\begin{table*}[t!]
\centering
\caption{Quantitative results of our method and baselines on the Area-4 of S3DIS dataset.}
\label{tab:s3dis_a4}
\resizebox{\textwidth}{!}
{
\begin{tabular}{crccccccccccccccc}
\toprule[1.0pt]
& & OA(\%) & mAcc(\%)& mIoU(\%) & ceil. & floor & wall & beam & col. & wind. & door & table & chair & sofa & book. & board. \\
\toprule[1.0pt]
\multirow{3}{*}{\makecell[l]{Supervised} }
& PointNet \cite{Qi2016} & 73.0 & 58.6 & 41.6 & 81.3 & 95.7 & 68.4 & 1.3 & 22.4 & 29.0 & 44.8 & 39.3 & 42.5 & 17.6 & 36.6 & 20.1\\

& PointNet++ \cite{Qi2017} & 74.8 & 66.4 & 47.7 & 85.5 & 96.1 & 69.9 & 4.4 & 23.8 & 27.0 & 50.5 & 44.9 & 54.0 & 35.6 & 38.4 & 43.8\\

& SparseConv \cite{Graham2018} & 88.3 & 76.2 & 65.5 & 93.0 & 94.9 & 78.2 & 53.3 & 57.9 & 43.4 & 59.1 & 69.4 & 76.6 & 55.1 & 73.8 & 30.9\\
\toprule[1.0pt]
\multirow{8}{*}{\makecell[c]{Unsupervised} } 
& Kmeans  & 17.9 & 19.9 & 7.8 & 18.6 & 18.2 & 10.6 & 0.9 & 3.8 & 5.2 & 11.7 & 5.8 & 7.4 & 2.4 & 8.7 & 0.4\\
& IIC \citep{Ji2019} & 33.0 & 13.5 & 8.2 & 14.9 & 25.9 & 35.1 & 0 & 1.1 & 1.4 & 3.7 & 4.1 & 0.9 & 0 & 10.8 & 0\\
& PICIE \citep{Cho2021} & 43.2 & 29.4 & 17.8 & 62.2 & 72.7 & 22.6 & 2.5 & 3.4 & 3.5 & 8.8 & 4.1 & 17.4 & 0 & 15.5 & 0.7\\
& GrowSP \citep{Zhang2023a} & 76.0 & \textbf{59.8} & 42.8 & 90.6 & 91.5 & 64.4 & \textbf{15.9} & \textbf{7.6} & 27.4 & 31.5 & 52.0 &\textbf{67.4} & 16.8 & 48.5 & 0\\
& PointDC \citep{Chen2023} & 54.0 & 35.3 & 25.2 & 87.9 & 86.8 & 24.7 & 0 & 4.2 & 12.2 & 18.7 & 32.6 & 17.6 & 1.9 & 15.6 & 0\\
& PointDC-DINOv2 \citep{Chen2023} & 73.0 & 46.5 & 39.8 & 89.4 & 91.9 & 58.2 & 0 & 0.7 & 25.3 &19.0 & 52.4 & 53.8 & 0.7 & 59.0 & 0\\
&\textbf{\nickname{} (Ours)}&\textbf{80.8} &54.4 & \textbf{43.5} & \textbf{92.5} & \textbf{93.8} & \textbf{73.9} & 4.1 & 0.1 & \textbf{34.8} & \textbf{44.6} & \textbf{54.0} & 64.9 & \textbf{69.2} & \textbf{73.4} & \textbf{33.5}\\
\bottomrule[1.0pt]
\end{tabular}
}
\end{table*}

\begin{table*}[t!]
\centering
\caption{Quantitative results of our method and baselines on the Area-5 of S3DIS dataset.}
\label{tab:s3dis_a5}
\resizebox{\textwidth}{!}
{
\begin{tabular}{crccccccccccccccc}
\toprule[1.0pt]
& & OA(\%) & mAcc(\%)& mIoU(\%) & ceil. & floor & wall & beam & col. & wind. & door & table & chair & sofa & book. & board. \\
\toprule[1.0pt]
\multirow{3}{*}{\makecell[l]{Supervised} }
& PointNet \cite{Qi2016} & 77.5 & 59.1 & 44.6 & 85.2 & 97.4 & 72.3 & 0.1 & 10.6 & 54.9 & 18.5 & 48.4 & 39.5 & 12.4 & 55.5 & 40.2\\

& PointNet++ \cite{Qi2017} & 77.5 & 62.6 & 50.1 & 83.1 & 97.2 & 66.4 & 0 & 8.1 & 55.6 & 15.2 & 60.4 & 64.5 & 36.6 & 58.3 & 55.7\\

& SparseConv \cite{Graham2018} & 88.4 & 69.2 & 60.8 & 92.6 & 95.9 & 77.2 & 0.1 & 36.7 & 37.6 & 59.8 & 77.2 & 83.9 & 59.7 & 78.5 & 30.4\\
\toprule[1.0pt]
\multirow{8}{*}{\makecell[c]{Unsupervised} } 
& Kmeans  & 21.4 & 21.2 & 8.7 & 18.7 & 18.0 & 16.7 & \textbf{0.2} & 2.5 & 12.0 & 5.7 & 8.7 & 5.6 & 0 & 13.6 & \textbf{2.3}\\
& IIC \citep{Ji2019} & 28.5 & 12.5& 6.4& 6.1 & 19.8 & 27.9 & 0 & 2.1 & 0.1 & 3.4 & 7.9 & 0.4 & 0 & 8.6 & 0\\
& PICIE \citep{Cho2021} &61.6 & 25.8 & 17.9 & 65.7 & 61.4 & 58.4 & 0 & 0.3 & 2.2 & 1.7 & 12.1 & 0 & 0 & 12.4 & 0\\
& GrowSP \citep{Zhang2023a} & 78.4 & \textbf{57.2} & 44.5 & 90.5 & 90.1 & 66.7 & 0 & \textbf{14.8} & 27.6 & \textbf{45.6} & 59.4 & 71.9 & \textbf{10.7} & 56.0 & 0.2\\
& PointDC \citep{Chen2023} & 55.5 & 35.1 & 23.9 & 84.4 & 84.3 & 30.2 & 0 & 1.8 & 12.2& 7.1 & 24.6 & 6.9 & 5.4 & 29.7 & 0.7\\
& PointDC-DINOv2 \citep{Chen2023} & 75.7 & 48.7 & 40.2 & 87.7 & 89.5 & 59.2 & 0 & 0.8 & 25.8 & 26.3 & \textbf{62.0} &68.3 &1.5 & 61.0 & 0.5\\
&\textbf{\nickname{} (Ours)}& \textbf{82.8} & 55.9 & \textbf{46.5} & \textbf{92.9} & \textbf{95.4} & \textbf{73.2} & 0 & 3.3 & \textbf{57.8} & 35.9 & 55.5 & \textbf{74.6} & 1.9 & \textbf{67.3} & 0.3\\
\bottomrule[1.0pt]
\end{tabular}
}
\end{table*}

\begin{table*}[t!]
\centering
\caption{Quantitative results of our method and baselines on the Area-6 of S3DIS dataset.}
\label{tab:s3dis_a6}
\resizebox{\textwidth}{!}
{
\begin{tabular}{crccccccccccccccc}
\toprule[1.0pt]
& & OA(\%) & mAcc(\%)& mIoU(\%) & ceil. & floor & wall & beam & col. & wind. & door & table & chair & sofa & book. & board. \\
\toprule[1.0pt]
\multirow{3}{*}{\makecell[l]{Supervised} }
& PointNet \cite{Qi2016} & 79.0 & 79.6 & 60.9 & 85.7 & 96.5 & 71.8 & 59.4 & 47.4 & 67.4 & 74.3 & 56.2 & 48.9 & 20.9 & 50.0 & 52.5\\

& PointNet++ \cite{Qi2017} & 82.0 & 89.3 & 69.0 & 87.5 & 96.3 & 76.8 & 66.4 & 54.4 & 72.1 & 77.4 & 64.3 & 66.5 & 43.7 & 51.8 & 70.2\\

& SparseConv \cite{Graham2018} & 91.6 & 87.3 & 80.5 & 97.4 & 95.0 & 83.4 & 83.0 & 75.1 & 81.1 & 74.9 & 81.3 & 84.3 & 79.0 & 80.7 & 61.4\\
\toprule[1.0pt]
\multirow{8}{*}{\makecell[c]{Unsupervised} } 
& Kmeans  & 21.0 & 25.0 & 10.4 & 18.6 & 17.6 & 8.9 & 11.3 & 0.6 & 14.8 & 17.6 & 12.0 & 8.7 & 0.3 & 7.8 &\textbf{6.2}\\
& IIC \citep{Ji2019} & 32.5 & 15.9 & 9.2 & 21.9 & 33.8 & 29.1 & 3.1 & 15.2 & 0 & 2.7 & 0.7 & 0 & 0 & 1.5 & 1.8\\
& PICIE \citep{Cho2021} & 39.3 & 28.5 & 17.8 & 56.9 & 61.7 & 18.6 & 20.5 & 4.2 & 6.0 & 8.7 & 14.7 & 15.9 & 1.1 & 5.7 & 0\\
& GrowSP \citep{Zhang2023a}& 75.6 & 58.5 & 47.6 & 89.4 & 88.0 & 57.7 & \textbf{70.6} & 2.0 & 32.4 & 36.7 & 63.2 & 69.8 & 1.5 & 58.9 & 0.2\\
& PointDC \citep{Chen2023} & 62.4 & 38.7 & 28.6 & 85.8 & 85.6& 43.8 & 5.0 & 16.5 & 8.8 & 10.7 & 41.7 & 12.5 & 0 & 33.0 & 0\\
& PointDC-DINOv2 \citep{Chen2023} &76.4 & 55.5 & 46.4 & \textbf{90.3} & 91.4 &61.7 & 0 & 19.7 & 63.1 &33.6 & 67.9 & 65.7 & 1.4 & 62.5 &0\\
&\textbf{\nickname{} (Ours)}&\textbf{77.9} & \textbf{62.9} & \textbf{50.6} &84.6 &\textbf{92.7} &\textbf{64.0} &25.7 &\textbf{23.9} &\textbf{65.9} &\textbf{38.8} &\textbf{68.9} &\textbf{72.2} &\textbf{2.5} &\textbf{68.4}  & 0\\
\bottomrule[1.0pt]
\end{tabular}
}
\end{table*}

\begin{table*}[t!]
\centering
\caption{Quantitative results of our method and baselines on the 6-fold validation on S3DIS dataset.}
\label{tab:s3dis_6fold}
\resizebox{\textwidth}{!}
{
\begin{tabular}{crccccccccccccccc}
\toprule[1.0pt]
& & OA(\%) & mAcc(\%)& mIoU(\%) & ceil. & floor & wall & beam & col. & wind. & door & table & chair & sofa & book. & board. \\
\toprule[1.0pt]
\multirow{3}{*}{\makecell[l]{Supervised} }
& PointNet \cite{Qi2016} & 75.9 & 67.1 & 49.4 & 85.0 & 94.5 &  68.9 & 30.0 & 23.6 & 50.1 &51.8 &44.3 & 48.7 & 18.1 &43.0 & 34.8\\

& PointNet++ \cite{Qi2017} & 77.1 & 74.1 & 55.1 & 85.7 & 91.6 & 69.8 & 36.0 & 28.0 &58.7 & 57.4 & 47.4 & 61.8 & 39.1 & 44.1 & 61.2\\

& SparseConv \cite{Graham2018} & 89.4 & 78.1 & 69.2 & 94.6 & 95.5 & 78.6 & 51.8 & 55.8 & 60.6 & 63.0 & 76.0 & 84.3 & 65.8 &  73.5 & 39.4\\
\toprule[1.0pt]
\multirow{8}{*}{\makecell[c]{Unsupervised} } 
& Kmeans  & 20.0 & 21.5 & 8.8 & 17.9 & 17.9 &11.6 & 6.5 & 1.8 & 8.2 & 11.1 & 7.9 & 7.9 & 1.5 & 8.4 & 4.8\\
& IIC \citep{Ji2019} & 32.8 & 14.7 & 8.5 & 18.9 & 30.0 & 29.3 & 1.7 & 7.2 & 0.4 & 3.4 & 2.8 & 2.5 & 0 & 5.5 & 0.6\\
& PICIE \citep{Cho2021} & 46.4 & 28.1 & 17.8 & 63.6 & 58.6 & 33.3 & 9.0 & 2.6 & 3.2 & 7.6 & 9.7 & 12.4 & 0.9 & 11.5 & 0.9\\
& GrowSP \citep{Zhang2023a}& 76.0 &\textbf{59.4} & 44.6 & \textbf{90.7} & 89.9 & 60.2 &\textbf{30.6} & \textbf{14.9} & 24.0 & 35.6 & 58.4 & \textbf{70.6} & 12.5 & 44.9 & 3.5\\
& PointDC \citep{Chen2023} &55.7 & 37.7 & 26.0 & 81.5 & 81.5 & 31.4 & 1.5 &  6.4 & 8.8 & 12.2 & 35.8 & 20.8 &2.6 & 27.5 &2.4\\
& PointDC-DINOv2 \citep{Chen2023} & 74.4 & 51.5 & 41.3 & 88.8 & \textbf{91.8} & 57.9 & 3.4 &10.4 & 27.5 & 28.3 & \textbf{61.3} & 64.8 & \textbf{13.8} &43.9 & 0.3\\
&\textbf{\nickname{} (Ours)} & \textbf{79.2} &58.0 & \textbf{46.3} & 90.2 & 89.5 & \textbf{68.7} & 13.4 &10.9 & \textbf{57.1} & \textbf{39.5} & 60.9 & 65.1 & 13.0 & \textbf{60.4}  &\textbf{5.7}\\
\bottomrule[1.0pt]
\end{tabular}
}
\end{table*}

\section{Evaluation on nuScenes}
\begin{figure*}[t!]
\centering
   \includegraphics[width=1\linewidth]{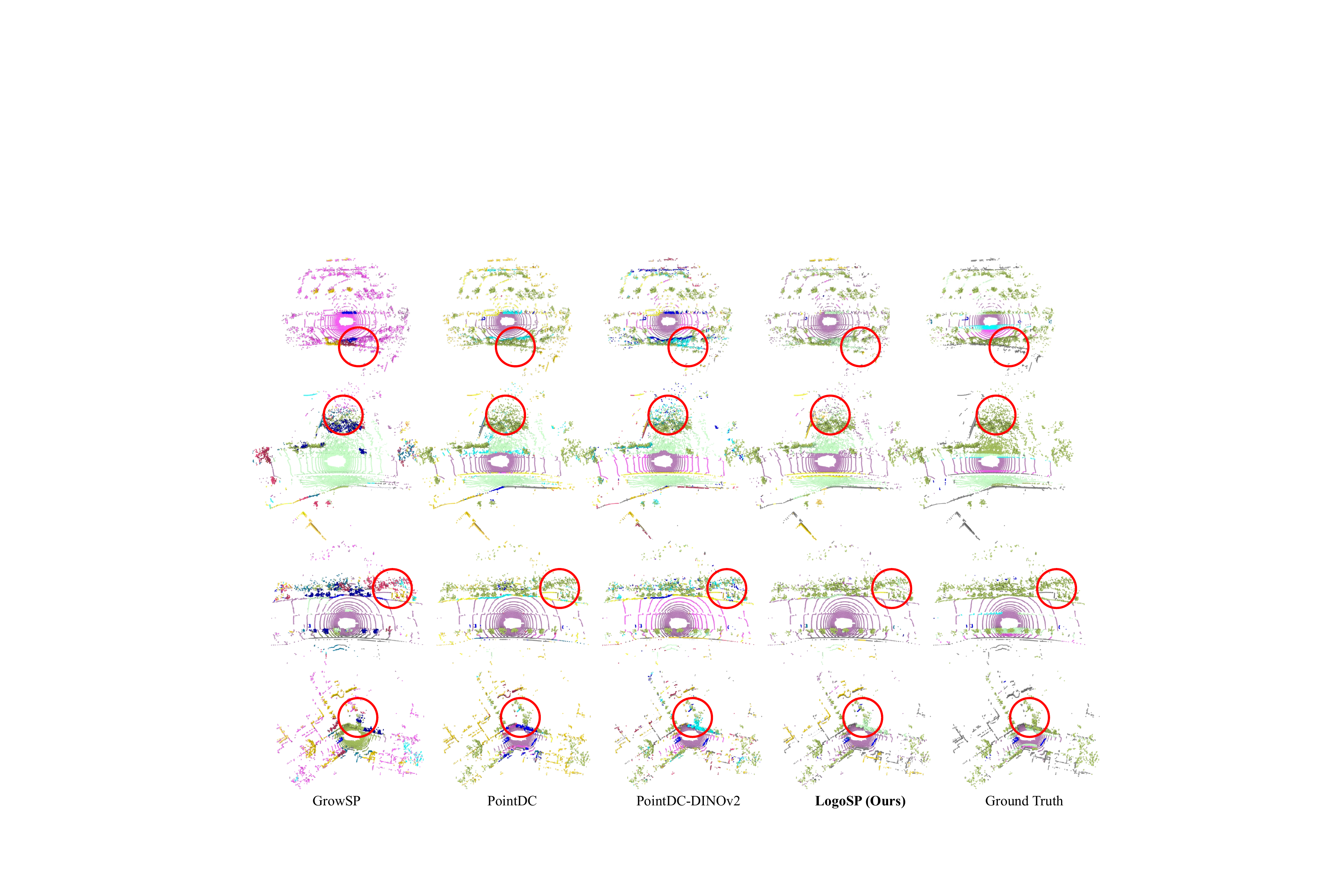}
\caption{Qualitative results of our method and baselines on the nuScenes dataset.}
\label{fig:app_nuscenes}
\end{figure*}

After obtaining initial superpoints using RANSAC and Euclidean Clustering, we employ a 15cm voxel grid to convert the point clouds into voxels for training the SparseConv backbone. We also utilize the ViT-S/14 configuration of DINOv2 and maintain the same distillation and segmentation training hyperparameters as used on ScanNet. The training set of nuScenes contains an extremely large number of point clouds, which are challenging to store in memory. Therefore, in each epoch, we randomly select 5,000 point clouds for training; this approach is also applied to all baseline models.

Table \ref{tab:nuscenes_val_detail} shows per-category results, where our model demonstrates superior performance on minor classes like \textit{truck} and \textit{car}. Qualitative results are shown in Figure \ref{fig:app_nuscenes}.

In addition to the experiments conducted on the validation set of nuScenes, we present segmentation results on its online hidden test set. This test set comprises 6008 outdoor point clouds categorized into 16 classes. Since there is no other unsupervised baseline evaluated on the hidden test set, we include successful fully-supervised methods for comparison. All models listed in Table \ref{tab:nuscene_test} are trained using the training set of nuScenes and then evaluated on the hidden test set. These results demonstrate the promising effectiveness of our unsupervised segmentation model.

\label{sec:app_unscenes}
\begin{table*}[t!]
\centering
\caption{Per-category quantitative results on the validation split of nuScenes dataset.}
\label{tab:nuscenes_val_detail}
\resizebox{\textwidth}{!}
{
\begin{tabular}{crccccccccccccccccccccccc}
\toprule[1.0pt]
& & OA(\%) & mAcc(\%)& mIoU(\%) & barrier. & bicycle. & bus. & car. & construction vehicle. & motorcycle. & pedestrian & traffic cone. & trailer. & truck. & drivable surface. & other flat. & sidewalk. & terrain. & manmade. & vegetation.\\
\toprule[1.0pt]
\multirow{8}{*} 
& GrowSP \citep{Zhang2023a}  &39.2 &17.5 &10.2 &7.5 &0 & 0.4 &42.9 &0.1 &0 &0.6 &0 &0.7 &1.4 &48.4 &\textbf{0.8} &6.5 &13.1 &21.4 &19.7\\

& PointDC \citep{Chen2023}&\textbf{56.8} &\textbf{29.4} &17.7 &11.6 &0 &0.5 &63.1 &\textbf{0.3} &0 &4.4 &0 &1.2 &26.4 &70.1 &0.1 &7.1 &19.3 &21.1 &\textbf{58.1} \\

& PointDC-DINOv2 \citep{Chen2023}&51.8 &28.6 &18.2 &\textbf{17.0} &0 &0.2 &58.4 &0.2 &0 &1.5 &0 &\textbf{1.6} &\textbf{43.3} &\textbf{71.8} &0 &\textbf{8.3} &\textbf{19.5} &17.6 &51.8 \\

&\textbf{\nickname{} (Ours)}&54.8 &29.2 &\textbf{20.1} &16.6 &0 &\textbf{0.7} &\textbf{70.2} &0.2  &\textbf{0.2} &\textbf{33.6}  &0  &0.3 &38.4 &59.4  &0.4  &8.0 &10.7 &\textbf{33.0} &49.3\\
\bottomrule[1.0pt]
\end{tabular}
}
\end{table*}

\begin{table*}
\centering
\caption{Per-category quantitative results on the \textbf{hidden test split} of nuScenes dataset.}
\label{tab:nuscene_test}
\resizebox{\textwidth}{!}
{
\begin{tabular}{crcccccccccccccccccccccccccccc}
\toprule[1.0pt]
&& mIoU(\%) & barrier. & bicycle. & bus. & car. & construction vehicle. & motorcycle. & pedestrain. & traffic cone. & trailer. & truck. & driveable. & other flat. & sidewalk. & terrain.& manmade. & vegetation.\\
\toprule[1.0pt]
\multirow{3}{*}{\makecell[c]{Supervised} }
& Cylinder3D \citep{Zhu2021b} &77.2 &82.8 &29.8 &84.3 &89.4 &63.0 &79.3 &77.2 &73.4 &84.6 &69.1 &97.7 &70.2 &80.3 &75.5 &90.4&87.6\\
& SPVNAS \citep{tang2020searching} &77.4 &80.0 &30.0 &91.9 &90.8 &64.7 &79.0 &75.6 &70.9 &81.0 &74.6 &97.4 &69.2 &80.0 &76.1 &89.3&87.1 \\
& Cylinder3D++ \citep{Zhu2021b}  &77.9 &82.8 &33.9 &84.3 &89.4 &69.6 &79.4 &77.3 &73.4 &84.6 &69.4 &97.7 &70.2 &80.3 &75.5 &90.4 &87.6\\
\toprule[1.0pt]
\multirow{1}{*}{\makecell[c]{Unsupervised} } 
& \textbf{\nickname{}(Ours)} &17.5 &12.4 &0 &1.9 &68.7 &0 &0.2 &22.5 &0 &0.1 &23.2 &62.8 &0.2 &1.4 &13.9 &30.1 &41.6 \\
\bottomrule[1.0pt]
\end{tabular}
}
\end{table*}